\documentclass[journal]{IEEEtran}
\usepackage{amsmath,amsfonts}
\usepackage{algorithmic}
\usepackage{algorithm}
\usepackage{array}
\usepackage[caption=false,font=normalsize,labelfont=sf,textfont=sf]{subfig}
\usepackage{textcomp}
\usepackage{stfloats}
\usepackage{verbatim}
\usepackage{graphicx}
\usepackage{float}
\usepackage{multirow}
\usepackage{cite}
\usepackage{url}
\usepackage[numbers,sort&compress]{natbib}
\usepackage{booktabs} % 专业表格线条
\usepackage{array}    % 增强列格式
\usepackage{xcolor}   % 可选：用于高亮行
\usepackage{placeins}
\usepackage{hyperref}

\begin{document}
\renewcommand\floatpagefraction{.9999}
\renewcommand\topfraction{.5}
\renewcommand\bottomfraction{.5}
\renewcommand\textfraction{.00001}
\setcounter{totalnumber}{10}
\setcounter{topnumber}{10}
\setcounter{bottomnumber}{10}
\renewcommand{\dblfloatpagefraction}{.3}
\title{MonoRelief V2: Leveraging Real Data for High-Fidelity Monocular Relief Recovery}

\author{Yu-Wei~Zhang, Tongju~Han, Lipeng~Gao, Mingqiang~Wei, Hui~Liu, Changbao~Li and Caiming~Zhang
	\thanks{ Yu-Wei Zhang, Tongju Han, Lipeng Gao and Changbao Li are with the School of Mechanical Engineering, Qilu University of Technology (Shandong Academy of Sciences) Jinan 250353, China. E-mail: zhangyuwei\_scott@126.com, tongju318@gmail.com, qluglp@163.com, 15550301926@163.com.}
	\thanks{Mingqiang Wei is with School of Computer Science and Technology, Nanjing University of Aeronautics and Astronautics, Nanjing 210016, China. Email: mqwei@nuaa.edu.cn}
	\thanks{Hui Liu is with School of Computer Science and Technology, Shandong University of Finance and Economics, Jinan 250014, China. E-mail: liuh\_lh@sdufe.edu.cn}
	\thanks{Caiming Zhang is with School of Computer Science and Technology, Shandong University, Jinan 250102, China. E-mail: czhang@sdu.edu.cn}}

% The paper headers
\markboth{Journal of \LaTeX\ Class Files,~Vol.~14, No.~8, August~2021}%
{Yu-Wei~Zhang,Tongju~Han,Lipeng~Gao,Changbao~Li,Hui~Liu and Caiming~Zhang,\MakeLowercase{\textit{et al.}}:MonoRelief V2: Leveraging Real Data for High-Fidelity Monocular Relief Recovery}

%\IEEEpubid{0000--0000/00\$00.00~\copyright~2021 IEEE}
% Remember, if you use this you must call \IEEEpubidadjcol in the second
% column for its text to clear the IEEEpubid mark.

\maketitle

\begin{abstract}
This paper presents MonoRelief V2, an end-to-end model designed for directly recovering 2.5D reliefs from single images under complex material and illumination variations. In contrast to its predecessor, MonoRelief V1 ~\cite{monorelief}, which was solely trained on synthetic data, MonoRelief V2 incorporates real data to achieve improved robustness, accuracy and efficiency. To overcome the challenge of acquiring large-scale real-world dataset, we generate approximately 15,000 pseudo-real images using a text-to-image generative model, and derive corresponding depth pseudo-labels through fusion of depth and normal predictions. Furthermore, we construct a small-scale real-world dataset (800 samples) via multi-view reconstruction and detail refinement. MonoRelief V2 is then progressively trained on the pseudo-real and real-world datasets. Comprehensive experiments demonstrate its state-of-the-art performance both in depth and normal predictions, highlighting its strong potential for a range of downstream applications.		Code is at: \href{https://github.com/glp1001/MonoreliefV2}{\texttt{https://github.com/glp1001/MonoreliefV2}}
\end{abstract}

\setcounter{figure}{0} %
\section{Introduction}
\IEEEPARstart{R}{elief}, a sculptural form that bridges the gap between 2D images and 3D objects, is prevalent in architecture, decorative arts, heritage preservation, and handicraft design. While Generative Artificial Intelligence has rapidly advanced in computer graphics and computer vision, with numerous text-to-image ~\cite{stable_diffusion,flux,midjourney} and text/image-to-3D ~\cite{tripo3d,meshy,hunyuan3d} models developed, dedicated research on generative 3D relief modeling remains scarce. One potential pipeline for this task could involve first generating a relief-style image from text or image prompts, and then converting it into a 3D relief model ready for CNC machining or 3D printing. Within this workflow, monocular relief recovery emerges as a critical and challenging step.

Reliefs exhibit diverse, material-dependent reflectivity (stone, metal, wood, et al.) and distinct shading effects, further modulated by lighting, texture, and thickness variations. Decoupling intrinsic geometry from these visual factors is inherently ill-posed. MonoRelief V1 ~\cite{monorelief} address this challenge by synthesizing large-scale relief images with ground-truth normal labels, fine-tuning Ominidata ~\cite{omnidata} for normal estimation, and reconstructing relief depths through depth-constrained normal integration (Fig. \ref{fig2}). While effective, this multi-stage approach suffers from limited generalization to real photographs and suboptimal computational efficiency.
\begin{figure*}[t] 
	\centering 
	{\tiny}	\includegraphics[width=7in]{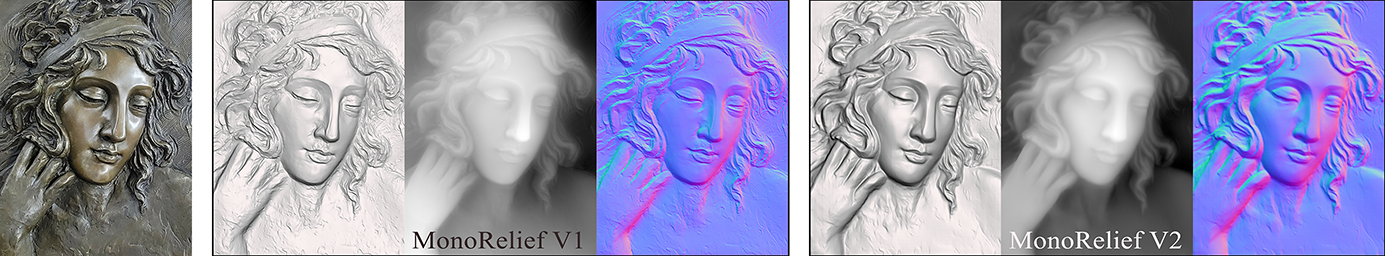} 
	\caption{MonoRelief V2 significantly outperforms V1 ~\cite{monorelief} in robustness, quality and efficiency.} 
	\label{fig1} 
\end{figure*}
\begin{figure}[b] 
	\centering 
	{\tiny}	\includegraphics[width=3.5in]{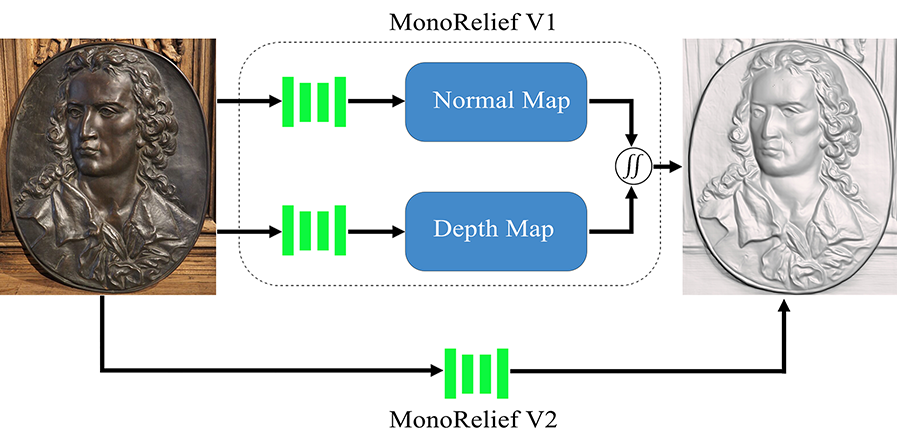} 
	\caption{MonoRelief V1 ~\cite{monorelief} versus V2. MonoRelief V2 directly recover reliefs from single images in an end-to-end manner, whereas MonoRelief V1 ~\cite{monorelief} relies on a three-stage pipeline, resulting in lower efficiency.} 
	\label{fig2} 
\end{figure}
In this work, we present MonoRelief V2, an end-to-end model trained on real relief data. Building such a dataset, however, is non-trivial: large-scale real-world relief images with accurate depth labels are costly and time-consuming to acquire, and no suitable public dataset currently meets these requirements. To address this gap, we propose a four-stage data construction and training strategy: (1) Leveraging the powerful text-to-image model Flux.1 ~\cite{flux}, we generate approximately ~15,000 pseudo-real relief images spanning diverse materials, objects, and thicknesses. Depth pseudo-labels are obtained by fusing predictions from DepthAnything V2 ~\cite{depthanything_v2} and MonoRelief V1 ~\cite{monorelief}. (2) We fine-tune the pre-trained DepthAnything V2 backbone ~\cite{depthanything_v2} on the pseudo-real dataset, yielding an initial model with strong generalization capability. (3) We capture multi-view photographs of physical reliefs under real-world conditions, and reconstruct accurate depths for 800 samples via multi-view reconstruction. (4) The initial model is fine-tuned on this small-scale real dataset using a lightweight adaptation strategy to enhance depth accuracy while maintaining generalization.

Through this progressive learning pipeline, MonoRelief V2 achieves state-of-the-art performance in depth and normal predictions, outperforming its predecessor V1 ~\cite{monorelief} and other leading methods across multiple metrics. Notably, the mean depth error drops from 21.766~\% (V1) to 14.025~\% on our new evaluation benchmark. Thanks to its robust generalization, high-fidelity reconstructions and efficiency, MonoRelief V2 establish a new paradigm for cost-effective, personalized relief customization and heritage preservation.

\section{Related works}
{\bf{Monocular Depth Estimation (MDE).}} Deep learning–based approaches have become dominant in MDE by effectively learning image representations from annotated data. In relative depth estimation, recent studies ~\cite{zeroshot_depth,recover_3d_shape,metric3d,zoedepth} have demonstrated remarkable zero-shot generalization by adopting multi-dataset joint training. MiDaS ~\cite{robust_monocular_depth} introduces an affine-invariant loss to address scale-variation problem across different datasets. DepthAnything ~\cite{depth_anything} uses large-scale unlabeled image collections to acquire additional visual priors, while its successor DepthAnything V2 ~\cite{depthanything_v2}, further adopts a teacher–student distillation framework to substantially improve prediction accuracy and robustness. Diffusion-based methods ~\cite{diffusion_depth,genpercept,lotus} have also emerged for dense vision tasks. Marigold ~\cite{diffusion_depth} adapts Stable Diffusion ~\cite{latent_diffusion} into a monocular depth estimator, alleviating the poor generalization and high data dependency of conventional approaches. Building on this work, GenPercept ~\cite{genpercept} and Lotus ~\cite{lotus} refine the framework to jointly predict relative depth maps and surface normals from a single image in a single forward inference.

In absolute depth estimation, Metric3D v2 ~\cite{metric3d_v2} introduces a canonical camera-space transformation to resolve depth ambiguities caused by varying camera intrinsics. UniDepth V1 ~\cite{unidepth} and UniDepth V2 ~\cite{unidepth_v2} jointly estimate camera intrinsics and integrate them into absolute depth prediction. DepthPro ~\cite{depth_pro} is first trained on a mixture of real and synthetic data, and then fine-tuned on synthetic dataset to enhance edge sharpness. MoGe V1 ~\cite{moge} predicts scale-invariant point maps, while its extension V2 ~\cite{moge_v2} augments the framework with a scale prediction head and depth-label optimization, further improving accuracy and edge sharpness. In contrast to these general-purpose methods, our work focuses specifically on the relief domain. We aim at developing an end-to-end absolute depth prediction model that generalizes across diverse shapes, lighting conditions and materials, while recovering both plausible depth structures and fine geometric details. A critical prerequisite for this goal is the construction of real relief dataset with broad data diversity.

{\bf{Monocular Normal Estimation.}} Surface normals encode each pixel’s local surface orientation, offering an intuitive representation of geometric details. Similar to monocular depth estimation, recent advances are dominated by deep learning–based approaches ~\cite{cross_task_consistency,aleatoric_uncertainty,3d_corruptions}. Omnidata ~\cite{omnidata} employs a Vision Transformer backbone with task-specific prediction heads and 3D data augmentation ~\cite{3d_corruptions} to enhance generalization. In contrast, DSINE ~\cite{rethink_inductive_biases} takes a data-efficient approach, introducing inductive biases to reduce learning complexity and achieving strong performance with only small-scale training sets. StableNormal ~\cite{stablenormal} adopts a diffusion-based architecture to firstly predict coarse normal maps, and then refine high-frequency details under semantic segmentation guidance. Marigold-E2E-FT ~\cite{finetune_diffusion} addresses the inference limitations of Marigold ~\cite{diffusion_depth} and fine-tunes its U-Net backbone, enabling superior zero-shot performance in a single deterministic pass. Unlike these methods, our goal is monocular depth recovery rather than direct normal prediction. We incorporate a normal-consistency loss during training to preserve geometric fidelity and fine surface details.
\begin{figure}[b] 
	\centering 
	{\tiny}	\includegraphics[width=3.5in]{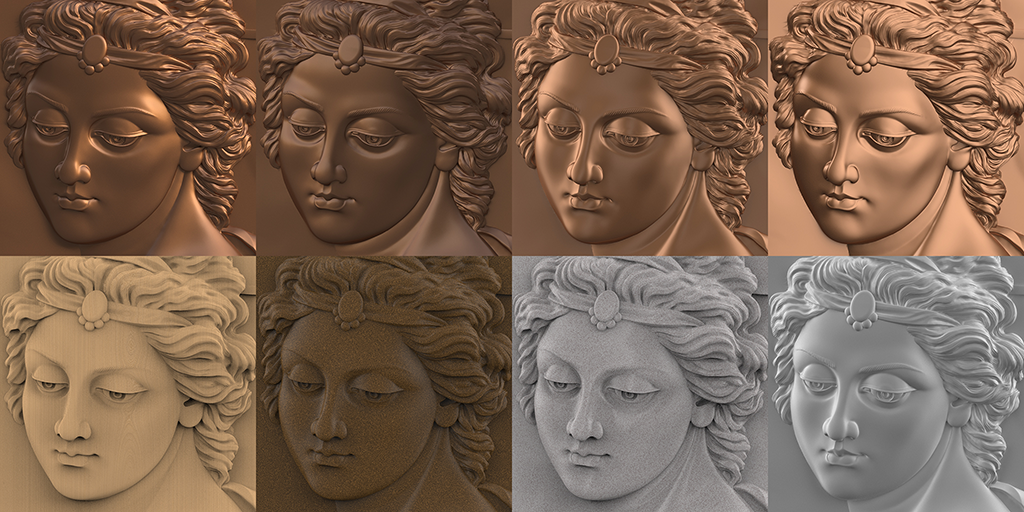} 
	\caption{Impact of physical properties on relief appearance. Top row: images rendered with identical geometry and material, but under variable lighting conditions. Bottom row: images rendered with identical geometry and lighting, but different materials and textures.} 
	\label{fig3} 
\end{figure}

\begin{figure}[t] 
	\centering 
	{\tiny}	\includegraphics[width=3.3in]{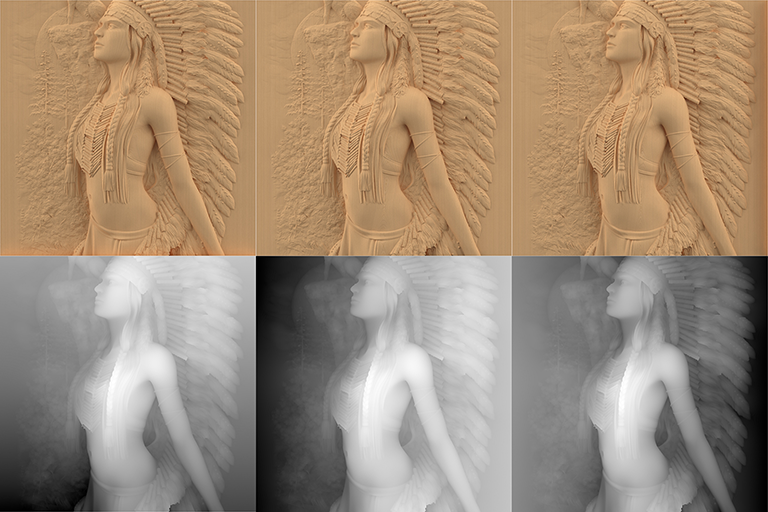} 
	\caption{Relief images and underlying depth maps. Under same material properties and illumination condition, reliefs with different depth structures might produce visually indistinguishable appearances.} 
	\label{fig4} 
\end{figure}
\begin{figure}[b] 
	\centering 
	{\tiny}	\includegraphics[width=3.3in]{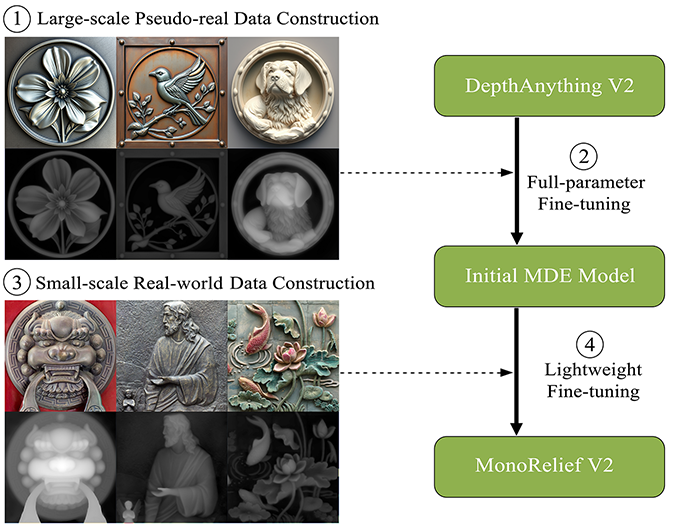} 
	\caption{Our four-stage framework.} 
	\label{fig5} 
\end{figure}

\begin{figure}[t] 
	\centering 
	{\tiny}	\includegraphics[width=3.3in]{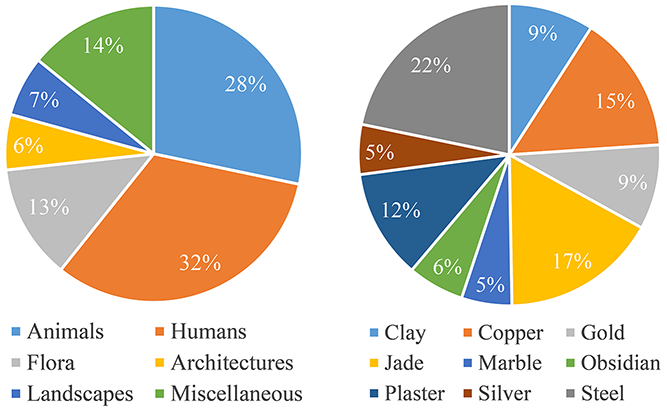} 
	\caption{Distribution of the pseudo-real images across object categories (left) and material types (right).} 
	\label{fig6} 
\end{figure}

\begin{figure*}[b] 
	\centering
	\includegraphics[width=5.5in]{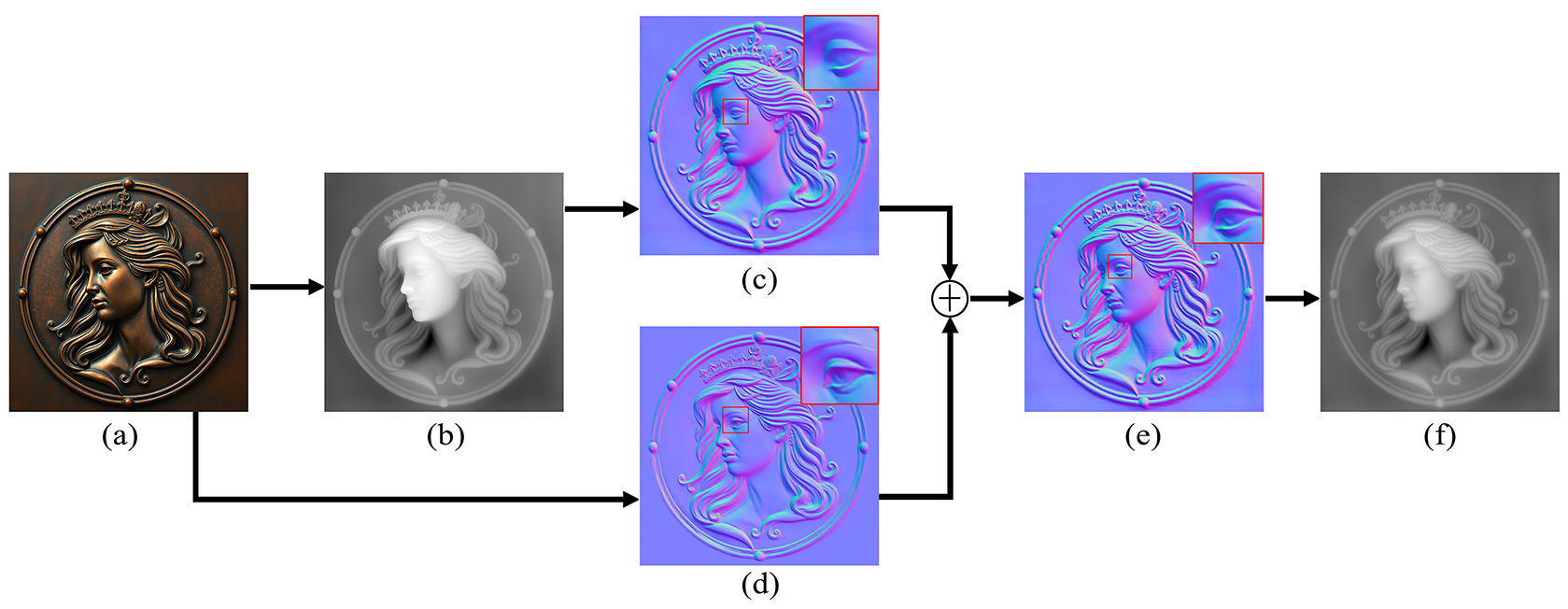} 
	\caption{Pipeline of depth pseudo-label generation. Given a pseudo-real relief image (a), a relative depth map (b) and a detailed normal map (d) are first predicted using the methods of ~\cite{depthanything_v2}  and ~\cite{monorelief} respectively. The relative depth map (b) is then globally scaled and gradient-transformed in the normal domain to produce an intermediate normal map (c). This map is subsequently fused with the detailed normal map (d) via soft blending, resulting in a refined normal map (e). Finally, iterative normal integration ~\cite{bilateral_normal} is applied to (e) to generate the final depth pseudo-label (f).} 
	\label{fig7} 
\end{figure*}

\begin{figure}[t] 
	\centering 
	{\tiny}	\includegraphics[width=3.2in]{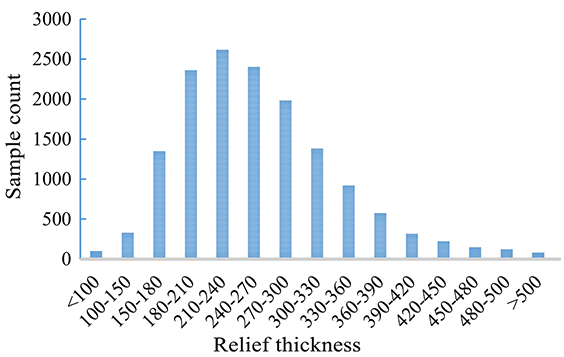} 
	\caption{Sample distribution across relief thicknesses in the pseudo-real dataset.} 
	\label{fig8} 
\end{figure}

\begin{figure*}[t] 
	\centering 
	\includegraphics[width=5.5in]{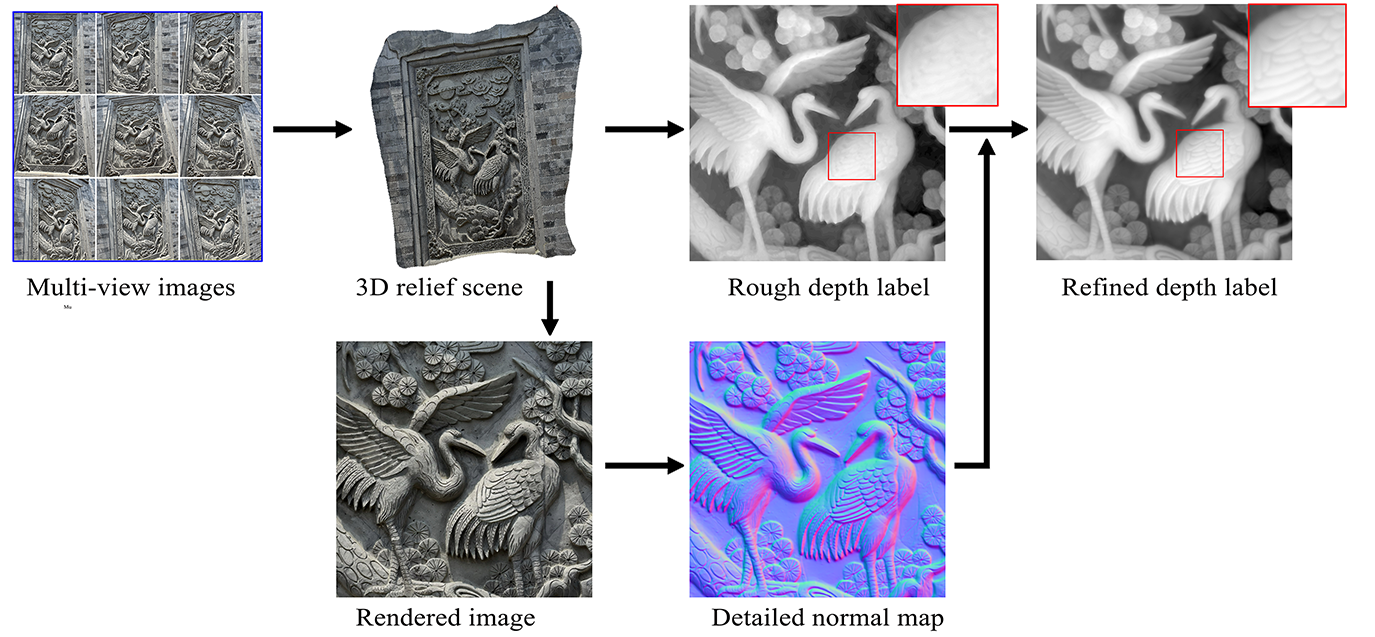} 
	\caption{Pipeline of real-world data construction. For each physical relief, we capture 20~30 multi-view images and reconstruct a textured 3D model using PolyCAM ~\cite{polycam} . From this model, we render an orthographic image and its rough depth label. To enhance label quality, we estimate a detail-rich normal map using the initial model and refine the rough depth label via depth-constrained normal integration.} 
	\label{fig9} 
\end{figure*}

{\bf{Relief Modeling.}} Relief modeling aims at transforming 3D models ~\cite{digital_basrelief,basrelief_generation,reliefnet} or 2D images ~\cite{portrait_relief,human_basrelief} into detail-rich reliefs via geometric transformation or depth estimation. Recent image-based methods often focus on specific object categories such as calligraphy ~\cite{calligraphy_relief}, flowers ~\cite{flower_relief}, portraits ~\cite{portrait_relief_v2,multistyle_relief}, or human bodies ~\cite{mmrelief}. In contrast, monocular relief recovery targets reconstructing accurate geometry from photographs of existing reliefs. MonoRelief V1 ~\cite{monorelief} pioneered this direction by generating large-scale synthetic relief datasets and fusing depth and normal predictions. In this work, we advance MonoRelief V1 by replacing synthetic training data with diverse real-relief datasets and adopting an end-to-end learning framework, thereby improving generalization, depth accuracy, and computational efficiency.

\section{Challenges of Monocular Relief Recovery}

\subsection{The Ill-Posed Nature of Monocular Geometry Decoupling.}
Relief sculpture is a distinct art form situated between 2D imagery and full 3D objects, fundamentally characterized by its shallow depth range and intricate surface details. The formation of a relief image arises from a coupled interplay between intrinsic geometry and extrinsic physical factors. Its visual appearance is jointly influenced by geometry, material reflectance, lighting conditions, and surface texture—any variation in these factors can significantly alter the final image, as shown in Fig. \ref{fig3}. For example, changes in lighting direction directly affect shading distribution and shadow formation, which can either conceal or emphasize geometric features. Specular highlights from glossy materials may distort perceived curvature, while the low-contrast reflectance of matte surfaces can suppress fine detail perception. Spatially varying albedo introduces geometry–texture ambiguity, where pixel intensity changes may originate from actual shape variation or merely from texture patterns alone. In addition, single-view imagery inherently suffers from depth ambiguity. As demonstrated in Fig. \ref{fig4}, reliefs with substantially different depth structure can produce nearly identical shading appearances. Consequently, monocular relief recovery—the task of decoupling relief geometry from a single image under unknown material, lighting, and texture—presents a highly ill-posed and under-constrained problem.

\subsection{Challenges of Real-world Relief Data Capture.}
Recent advances in monocular depth estimation ~\cite{metric3d_v2,unidepth,unidepth_v2,depth_pro,moge,moge_v2} have shown that deep neural networks trained on large-scale and diverse real-world datasets can achieve high prediction accuracy. Extending this paradigm to relief recovery is theoretically attractive but fundamentally depends on access to a large-scale, high-quality, and diverse dataset containing paired relief images and depth labels. While high-fidelity photographs of reliefs can be found online, many publicly accessible examples are synthetic renderings created with tools such as ZBrush or 3ds Max, which exhibit substantial domain gaps compared to real-world imagery. Commercial repositories (e.g., CGTrader ~\cite{cgtrader}, Zeel ~\cite{Zeel}, Sketchfab ~\cite{Sketchfab}) supply artist-designed 3D assets, but these often lack realistic texture maps and physically accurate material definitions. While acquiring paired data via 3D scanning or multi-view reconstruction is theoretically possible, yet in practice it is hindered by high equipment costs and the substantial labor required for data capture and post-processing.

\begin{figure*}[b] 
	\centering 
	\includegraphics[width=6.5in]{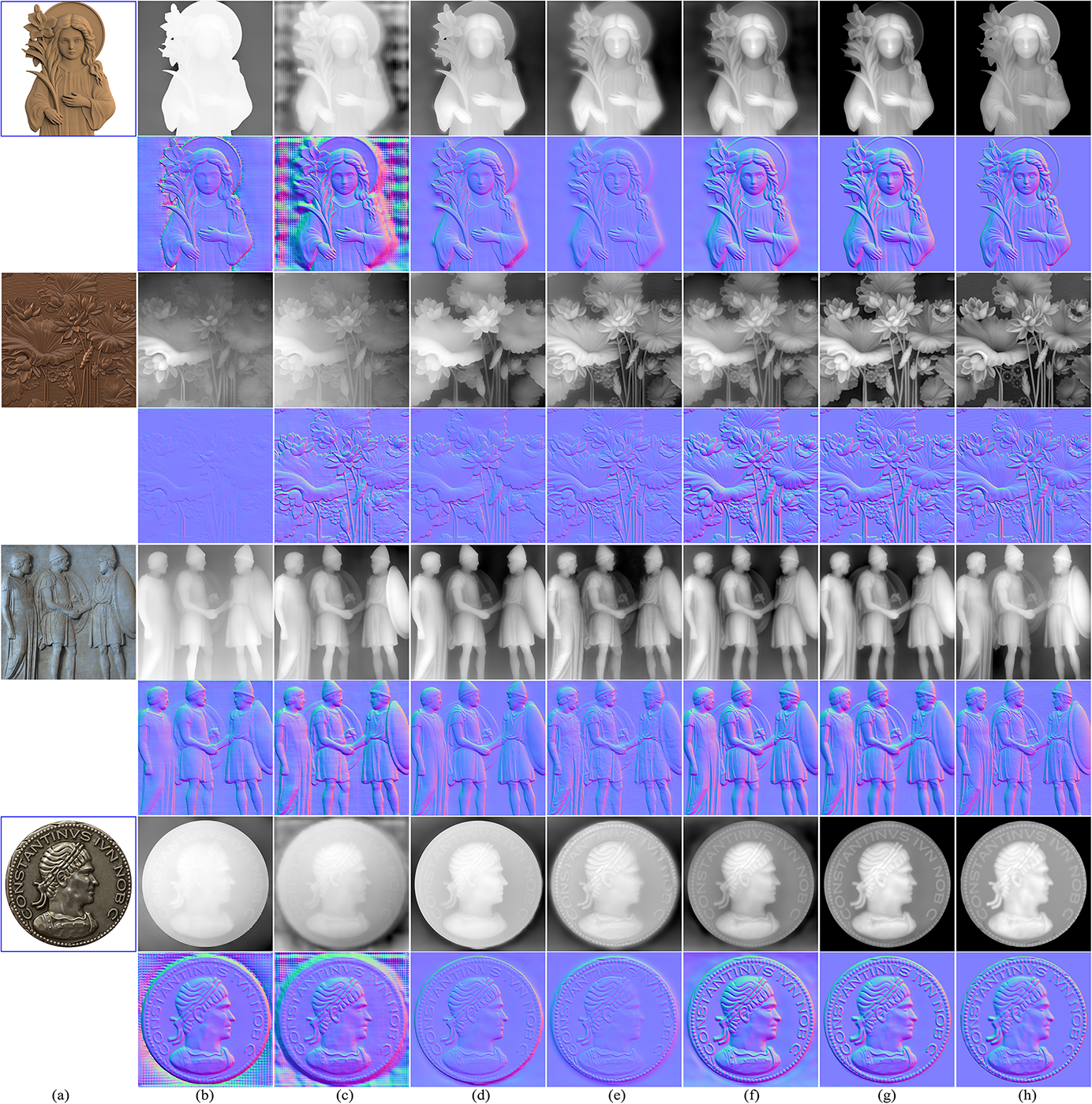} 
	\caption{Comparison of different methods on depth prediction (normalized to [0, 255] range). (a) input images (b) DepthPro ~\cite{depth_pro} (c) MoGe V2 ~\cite{moge_v2} (d) DepthAnything V2 ~\cite{depthanything_v2} (e) MonoRelief V1 ~\cite{monorelief} (f) MonoRelief V2-initial model (g) MonoRelief V2-final model (h) ground-truth. Best viewed with zoom.} 
	\label{fig10} 
\end{figure*}

\begin{figure*}[b] 
	\centering 
	{\tiny}	\includegraphics[width=6.6in]{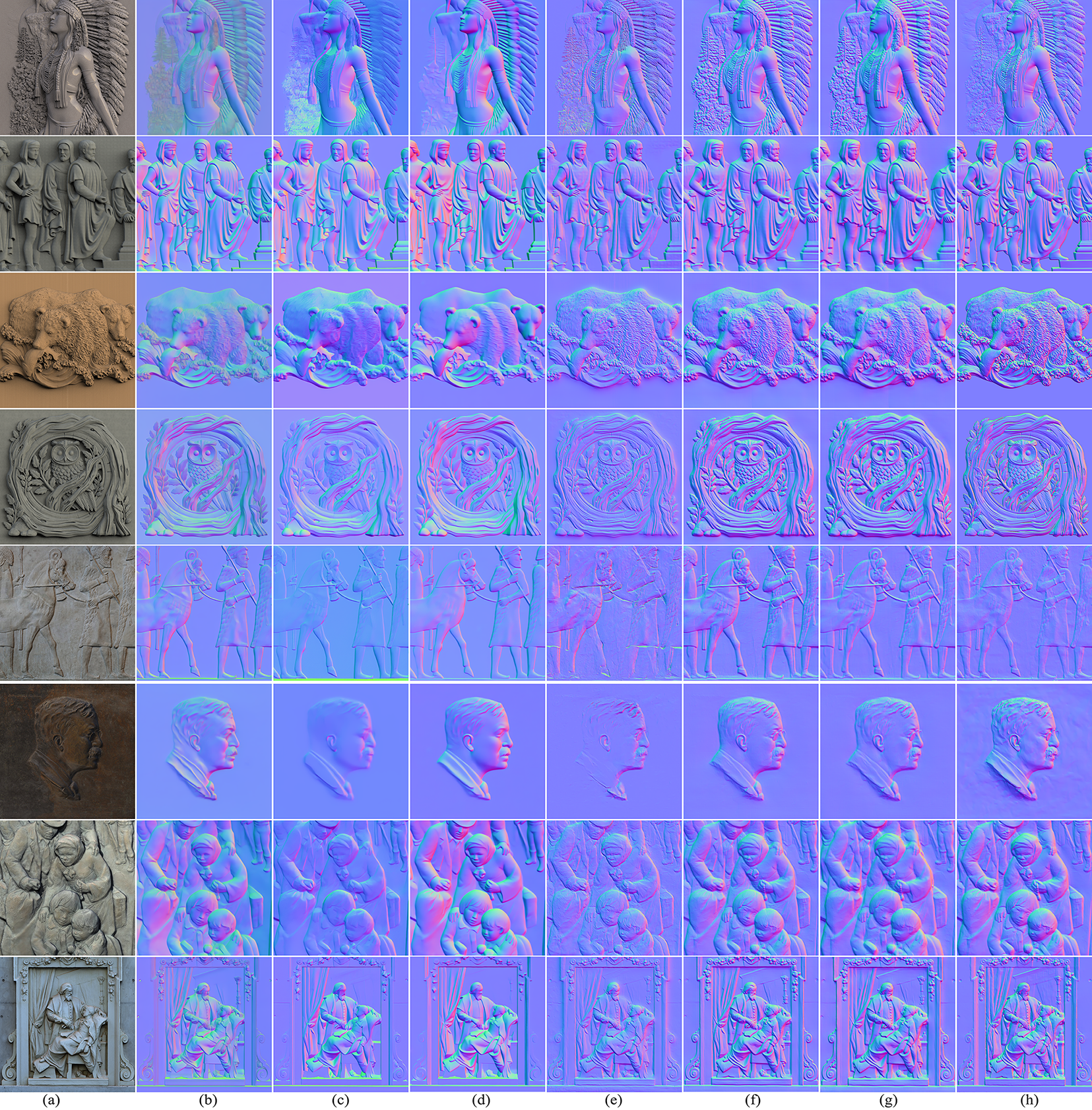} 
	\caption{Comparison of different methods on normal prediction. (a) input images (b) Lotus ~\cite{lotus} (c) StableNormal ~\cite{stablenormal} (d) Marigold-E2E-FT ~\cite{finetune_diffusion}(e) MonoRelief V1-normal ~\cite{monorelief} (f) MonoRelief V2-initial model (g) MonoRelief V2-final model (h) ground-truth. Best viewed with zoom.} 
	\label{fig11} 
\end{figure*}

\begin{figure*}[b] 
	\centering 
	{\tiny}	\includegraphics[width=6.6in]{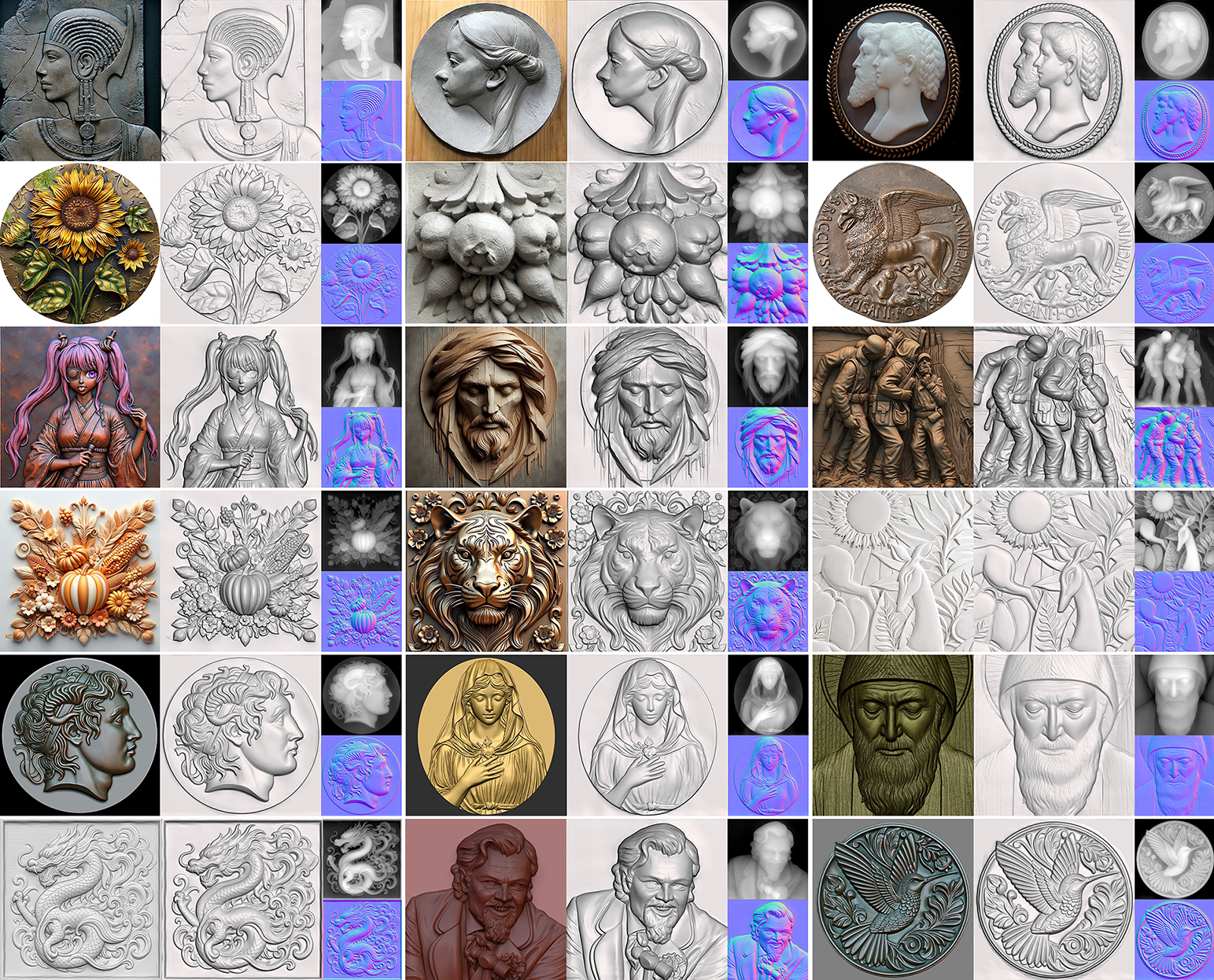} 
	\caption{Results of MonoRelief V2 on real-world relief photographs (top two rows), AI-generated images (middle two rows), and software-rendered graphics (last two rows). For each recovered relief, we show its MeshMixer rendering, normalized depth map, and normal map respectively.} 
	\label{fig12} 
\end{figure*}

\section{MonoRelief V2}
\subsection{Overview}
Given the difficulty of acquiring large-scale real-world data, we propose a novel dataset construction framework coupled with a progressive fine-tuning strategy. Our approach consists of four stages, as illustrated in Fig. \ref{fig5}.
\begin{enumerate}
\item{Pseudo-real dataset construction: Pseudo-real relief images are synthesized using a text-to-image generative model, and corresponding depth pseudo-labels are generated through a depth-normal fusion process.}
\item{Initial model training: An initial depth estimation model is trained by fully fine-tuning DepthAnything V2 ~\cite{depthanything_v2} on the pseudo-real dataset.}
\item{Real-world dataset construction: A small-scale real-world dataset is constructed by capturing multi-view images of physical reliefs, followed by multi-view 3D reconstruction and geometric detail refinement to produce paired images and depth labels.}
\item{Lightweight fine-tuning:  We apply lightweight fine-tuning of the initial model on the real-world dataset, enhancing depth estimation accuracy while maintaining generalization capability.}
\end{enumerate}

\subsection{Pseudo-real dataset construction}
{\bf{Pseudo-real Image Generation.}} We utilize the open-source text-to-image model Flux.1-schnell ~\cite{flux} to generate pseudo-real relief images. To ensure diversity in both semantic content and visual appearance, we devise a two-level prompting scheme:
\begin{enumerate}
\item{Object-category prompts: Covering common relief themes, including animals (e.g., canine, feline, elephantine, bovine), humans (full-body or half-body, single or multiple subjects), flora, architectures, landscapes, and a miscellaneous category (empty prompts).}
\item{Material-attribute prompts: Incorporating nine representative relief materials—clay, copper, gold, jade, marble, obsidian, plaster, silver, and steel—to simulate varied reflections and shading characteristics.}
\end{enumerate}

Using the standardized template ``A/an $<$material$>$ relief of a/an $<$object$>$'', we generated images at 1024×1024 resolution with four denoising steps and random seed sampling. This process initially yielded approximately 30,000 candidates. Owning to Flux.1-schnell’s limitations, some outputs exhibited shape distortions or insufficient high-frequency detail. To ensure data quality, we conducted a dual-blind review by two domain experts, focusing on shape and detail fidelity. Through this process, a curated subset of 15,000 high-quality pseudo-real images were selected. These images exhibit natural shading appearances and rich surface details (see Fig. \ref{fig15} in Appendix). Fig. \ref{fig6} presents the distribution of samples across object categories and material types. Despite being labeled as “pseudo-real”, many of these generative images are visually indistinguishable from photographs of real-world reliefs.

{\bf{Depth Pseudo-label Generation.}} Following MonoRelief V1 ~\cite{monorelief} and MMRelief ~\cite{mmrelief}, we adopt a depth-normal fusion strategy to generate depth pseudo-labels. Monocular depth estimation typically provides accurate occlusion boundaries and plausible coarse depth structures, whereas monocular normal estimation excels at capturing high-frequency geometric details. However, direct applying depth-constrained normal integration ~\cite{monorelief,mmrelief} can introduce artifacts in regions where these two modalities are inconsistent. To address this, we propose a novel fusion pipeline (Fig. \ref{fig7}), as follows:
\begin{enumerate}
	\item{{\bf{Depth and Normal Prediction.}} We first predict a relative depth map using DepthAnything v2 ~\cite{depthanything_v2} and a detailed normal map using MonoRelief V1 ~\cite{monorelief}. The relative depths are converted to absolute depths via a normal-guided global scaling operation, as in ~\cite{monorelief}.}
	\item{{\bf{Normal Transformation.}} A nonlinear normal transformation from ~\cite{monorelief} is applied to the surface normals derived from the scaled absolute depth. This suppresses large gradients at occlusion boundaries while preserving subtle ones in smoothly varying regions.}
	\item{{\bf{Geometric Fusion.}} We blend the transformed normal map $N_1$ with the detailed normal map $N_2$ via soft fusing scheme:	
{\scriptsize
	\begin{equation}
		N^* = 
		\begin{cases} 
			N_1 - (1 - 2N_2) \cdot N_1 \cdot (1 - N_1) & \text{if } N_2 \leq 0.5, \\
			N_1 + (2N_2 - 1) \cdot (\sqrt{N_1} - N_1) & \text{if } N_2 > 0.5.
		\end{cases}\label{eq1}
	\end{equation}
}	

	This formulation mitigates local inconsistencies, transfers fine details from $N_2$ into $N_1$, and preserves the global hierarchical structure of $N_1$.}	
	\item{{\bf{Depth Label Generation.}} The fused normal map is vector‐normalized and integrated into a depth map using an iterative normal integration process ~\cite{bilateral_normal}. Compared to a single-pass integration ~\cite{monorelief,mmrelief}, multiple iterations sharpen occlusion edges and produce more accurate depth labels.}
\end{enumerate}	

Through the above pipeline, we generated a depth pseudo-label for each pseudo-real image. As illustrated in Fig. \ref{fig8}, the resulting relief thicknesses distribution follows a Gaussian-like curve, spanning from shallow bas-reliefs to prominent high-reliefs.

\subsection{Initial model training}
DepthAnything V2 ~\cite{depthanything_v2}  is among the leading MDE models in terms of both generalization and efficiency. Its encoder is a DINO V2-pretrained Vision Transformer (ViT-small/base/large) that extracts multi-scale features, while its decoder follows the Dense Prediction Transformer (DPT) design to produce pixel-wise depth maps.

To leverage the rich semantic priors of DepthAnything V2 ~\cite{depthanything_v2} , we fully fine-tuned its ViT-base variant on our pseudo-real relief dataset. The model was trained using a composite loss function that combines absolute-depth regression and surface-normal alignment:
\begin{equation}
	\text{Loss} = \frac{1}{M} \sum_{i=1}^{M} \left[ \alpha \cdot \| \mathbf{d}_i^{\text{pred}} - \mathbf{d}_i^{\text{gt}} \| - \langle \mathbf{n}_i^{\text{pred}} \cdot \mathbf{n}_i^{\text{gt}} \rangle \right]\label{eq2}
\end{equation}
where $M$ denotes the number of pixels in an image, $d_i^{\text{pred}}$ and $d_i^{\text{gt}}$ are the predicted and ground-truth depths, and $n_i^{\text{pred}}$ and $n_i^{\text{gt}}$ are normal vectors computed from their respective depth maps. The weight $\alpha$ was scheduled during training. It was set to 0.1 for the first ten epochs to priorities depth accuracy, and then reduced to 0.01 for the remaining epochs to suppress geometric noise. Training was conducted for 20 epochs on a single NVIDIA A100, with a learning rate of 5 × $10^{-5}$ and a batch size of 4.

\subsection{Real-world Dataset Construction}
Although the fully fine-tuned initial model exhibits strong generalization and recovers reliefs with coherent depth structures and rich geometric details, residual inaccuracies in the depth pseudo-labels introduce noticeable errors during supervised training. These inaccuracies primarily arise from the structural distortions in DepthAnything V2's depth predictions and global orientation biases in MonoRelief V1's normal estimations.

To enhance depth prediction accuracy, we constructed a compact real-world relief dataset with reliable depth labels. We documented 800 physical reliefs—crafted from diverse materials including gypsum, masonry, metal, and wood—across parks, museums, and heritage sites (see Fig. \ref{fig16} in the Appendix). For each relief, we captured 20–30 high-resolution images from multiple viewpoints and performed multi-view 3D reconstruction using Polycam ~\cite{polycam} . The resulting texture-aligned 3D meshes were imported into KeyShot ~\cite{keyshot}  for orthographic rendering and depth sampling. Because Polycam ~\cite{polycam}  reconstructions often lack fine geometric details, we refined the raw depth labels via depth-constrained normal integration. First, we estimated a detail-rich normal map using our initial model. Then, we optimized the depth label by minimizing:
\begin{equation}
	\min_{z} \iint_{\Omega} \left( \nabla z_i - g_i \right)^2 + \mu \cdot \left( z_i - d_i \right)^2 \label{eq3}
\end{equation} 
where $\Omega$ denotes the image domain, $d_i$ is the initial depth label, and $g_i$ represents the gradient field derived from the estimated normals. The first term injects high-frequency details from normal map, while the second term ($\mu$=0.02) preserve the global depth structure. 
Fig. \ref{fig9} illustrates the complete pipeline. Statistical analysis reveals that the refined depth labels (each at 1024×1024 pixels) span a depth range of 12.81-445.73 pixels, closely matching the thickness distribution of the pseudo-real dataset.

\subsection{Lightweight Adaptation}
While the real-world relief dataset introduced in Section 4-D provides higher-precision depth labels than the pseudo-real dataset, its limited scale makes full-parameter fine-tuning prone to overfitting. Such overfitting risks eroding the valuable priors learned from the pseudo-real data, ultimately degrading the model’s generalization ability. To address this, we adopt a Low-Rank Adaptation (LoRA) strategy on the initial model, which freezes the majority of parameters and fine-tunes only a small set of trainable weights, enabling depth refinement without overwriting the initial model’s knowledge.

Following MonoRelief V1 ~\cite{monorelief}, we inserted LoRA adapters into the projection matrices of the self-attention modules and the fully connected layers within the MLP blocks, using a rank of 8. This configuration restricts trainable parameters to just $1.196\%$ of the total. We optimized the model using the same composite loss from Eq. \eqref{eq2} (an absolute depth loss term and a normal angle loss term), for 20 epochs with a learning rate of 5 × $10^{-5}$ and a batch size of 4.

We also evaluated an alternative adaptation strategy by freezing the entire ViT-base encoder and fine-tuning only the last five decoder layers. Although this strategy was able to reduce the depth error, it compromised generalization and produced blurred geometric details. In contrast, the lightweight LoRA-based adaptation not only preserves strong generalization, but also achieves a larger decrease in depth error while maintaining the model’s ability to recover fine geometric details.

\begin{table*}[t]
	\centering
	\caption{Depth Prediction Evaluation}
	\label{tab1}
	\renewcommand{\arraystretch}{1} % 调整行高
	\begin{tabular}{@{}|l|c|c|c|c|c|c|c|@{}} % @{} 去除左右边距
		\hline
		Method & 
		\boldmath$\varepsilon_d \downarrow$ & 
		Depth PSNR $\uparrow$ & 
		Depth SSIM $\uparrow$ & 
		\boldmath$\varepsilon_n \downarrow$ & 
		Normal PSNR $\uparrow$ & 
		Normal SSIM $\uparrow$ & 
		Ranking \\
		\hline
		DA V2 \cite{depthanything_v2}      & 16.415 & 14.575 & 0.715 & 15.654 & 21.855 & 0.798 & 3.166 \\ \hline
		DepthPro \cite{depth_pro}  & 73.558 & 11.378 & 0.625 & 23.674 & 21.149 & 0.733 & 6.0   \\ \hline
		MoGe V2 \cite{moge_v2}   & 68.961 & 12.823 & 0.693 & 16.604 & 21.344 & 0.766 & 5.0   \\ \hline
		MonoRelief V1 \cite{monorelief} & 21.766 & 14.840 & 0.712 & 16.213 & 21.548 & 0.783 & 3.833 \\ \hline
		Our initial model     & 19.766 & 16.256 & 0.742 & 14.441 & 22.582 & 0.820 & 2.0   \\ \hline
		Our final model       & \textbf{14.025} & \textbf{17.739} & \textbf{0.789} & \textbf{14.327} & \textbf{22.602} & \textbf{0.821} & 1.0 \\ \hline
	\end{tabular}
\end{table*}

\section{Experiments}
\subsection{Benchmark}
We constructed a new benchmark for relief recovery evaluation comprising of both synthetic and real-world data. The synthetic subset contains 252 samples generated by rendering 36 hand-crafted relief models in KeyShot ~\cite{keyshot} . Each relief was rendered with seven distinct material types—leather, concrete, high-gloss lacquer, bamboo, silver, marble, and porcelain—and illuminated under lighting conditions randomly sampled from ten HDRI environments (five indoor and five outdoor). The real-world subset consists of 41 physical relief models, with predefined material properties and texture maps sourced from CGTrader ~\cite{cgtrader} , Zeel ~\cite{Zeel} , and Sketchfab ~\cite{Sketchfab} . These models was rendered in Keyshot ~\cite{keyshot}  under six different lighting setups to yield a diverse range of appearances. For models exhibiting rich geometrical detail, labels were obtained via direct depth sampling. For those lacking fine geometry, we further refined the depth labels using the method described in Section 4-D (Eq. \eqref{eq3}). In total, the benchmark comprises 498 samples at a resolution of 1024×1024, with depth ranges spanning from 27.78 to 296.36 pixels.

\subsection{Depth Prediction Evaluation}
We evaluate MonoRelief V2 on this benchmark against MonoRelief V1 ~\cite{monorelief} and other leading MDE models, including DepthAnything V2 ~\cite{depthanything_v2} , DepthPro ~\cite{depth_pro} , and MoGe V2 ~\cite{moge_v2} . The evaluation metrics are:
\begin{enumerate}
	\item{{\bf{Mean depth error percentage $\varepsilon_d$:}} 
		
		$\varepsilon_d = \frac{1}{N} \sum_{j=1}^{N} \left[ \frac{1}{M} \sum_{i=1}^{M} \frac{\left| d_i - d_i^* \right|}{\max \left( d_i^* \right)} \right]$ where $M$ and $N$  denotes the number of pixels per image and the number of testing images, respectively. $d_i$ and $d_i^*$ are the predicted and ground-truth depths at pixel i, and max($d_i^*$) is the maximum ground-truth depth in that image.}
	\item{{\bf{Depth map PSNR and SSIM:}}
		 
		These metrics assess the pixel-wise fidelity and structural similarity between the predicted and ground-truth depth maps, both normalized to [0, 255] range.}
	\item{{\bf{Normal angular error $\varepsilon_n$:}}
		
		 $\varepsilon_n = \frac{1}{N} \sum_{j=1}^{N} \left[ \frac{1}{M} \sum_{i=1}^{M} \arccos \left( \mathbf{n}_i \cdot \mathbf{n}_i^* \right) \right]$, where $n_i$ and$ n_i^*$ are the predicted and ground-truth surface normal at pixel i, derived from their respective depth maps.}
	\item{{\bf{Normal map PSNR and SSIM:}} 
		
		These metrics evaluate the fidelity and structural consistency of the predicted normal maps relative to the ground-truth, focusing on geometrical details.}
\end{enumerate}

As shown in Table \ref{tab1}, MonoRelief V2 surpasses all competing methods across every evaluation metric, reflecting a comprehensive improvement in relief recovery. Compared to MonoRelief V1 ~\cite{monorelief}, its mean depth error $\varepsilon_d$ drops from 21.7666~\% to 14.0256~\%, and its mean normal angular error $\varepsilon_n$ falls from 16.213 to 14.327 . Even when trained solely on pseudo-real data, the V2 initial model achieves the second-best overall performance. Lightweight fine-tuning on real-world data further reduces $\varepsilon_d$ from 19.7666~\% to 14.0256~\%, and $\varepsilon_n$ from 14.441 to 14.327, highlighting the importance of real-data adaptation. In contrast, DepthAnything V2 ~\cite{depthanything_v2} , DepthPro ~\cite{depth_pro} , and MoGe V2 ~\cite{moge_v2}  underperform due to their design for general scenes and lack of relief-specific training data. Consequently, they struggle with diverse material properties and complex shading variations.

\begin{table*}[t]
	\centering
	\caption{Normal Prediction Evaluation}
	\label{tab2}
	\begin{tabular}{|l|c|c|c|c|c|c|}
		\hline
		Methods & Mean normal angular error (↓) & 11.25°(↑) & 22.5°(↑) & Normal PSNR (↑) & Normal SSIM (↑) & Ranking \\ \hline
		GenPercept [18] & 29.033 & 9.831 & 31.961 & 20.583 & 0.788 & 6.6 \\ \hline
		DSINE [29] & 22.919 & 17.545 & 58.818 & 18.387 & 0.779 & 6.4 \\ \hline
		Lotus [19] & 27.374 & 11.987 & 39.636 & 20.221 & 0.793 & 5.8 \\ \hline
		StableNormal [30] & 21.145 & 30.116 & 76.129 & 20.147 & 0.767 & 5.6 \\ \hline
		Marigold-E2E-FT [31] & 17.685 & 38.363 & 76.909 & 19.532 & 0.788 & 4.4 \\ \hline
		MonoRelief V1-Normal [1] & 17.510 & 41.181 & 73.792 & 20.816 & 0.773 & 4.2 \\ \hline
		Our initial model & 14.441 & 51.064 & 86.532 & 22.582 & 0.820 & 1.8 \\ \hline
		Our final model & \textbf{14.327} & \textbf{52.207} & 85.714 & \textbf{22.602} & \textbf{0.821} & \textbf{1.2} \\ \hline
	\end{tabular}
\end{table*}

Fig. \ref{fig10} illustrates that both our initial and final models significantly outperform the baselines in depth structure similarity and normal map fidelity. The initial model’s residual depth errors mainly concentrate in background planar regions, a byproduct of using DepthAnything V2 ~\cite{depthanything_v2}  predictions as depth constraints during pseudo-real data construction, which introduces errors into the depth labels. After adaptation, the final V2 model substantially improves accuracy in these planar regions and reduces overall depth error. By comparison, DepthAnything V2 ~\cite{depthanything_v2}  lacks geometric detail, DepthPro ~\cite{depth_pro}  and MoGe V2 ~\cite{moge_v2}  exhibit limited generalization on relief images, and MonoRelief V1 ~\cite{monorelief} suffers from background-plane artifacts and large depth inaccuracies.

\subsection{Normal Prediction Evaluation}
We further evaluate the normal predictions of MonoRelief V2 by comparing it against leading monocular normal estimation methods, including DSINE ~\cite{rethink_inductive_biases} , GenPercept ~\cite{genpercept} , Lotus ~\cite{lotus} , Stable Normal ~\cite{stablenormal} , Marigold-E2E-FT ~\cite{finetune_diffusion}, and MonoRelief V1-Normal ~\cite{monorelief}. The evaluation metrics include mean normal angular error, proportions of pixels with angular error below 11.25° and 22.5°, as well as normal map PSNR/SSIM. As shown in Table \ref{tab2}, both our initial and final models significantly outperform all baselines across every metric, establishing a new state-of-the-art. Notably, the final model achieves a mean angular error of just 14.327°.

Qualitative results in Fig. \ref{fig11} confirm these improvement across diverse relief images featuring various materials and lighting conditions. Both versions of MonoRelief V2 produce normal maps that closely align with the ground truth while preserving rich geometric detail. In contrast, MonoRelief V1-normal ~\cite{monorelief}, trained on synthetic data, often produces structural distortions and geometric noise when applied to real-world reliefs. Diffusion-based methods like Lotus ~\cite{lotus} , StableNormal ~\cite{stablenormal} , and Marigold-E2E-FT ~\cite{finetune_diffusion} yield sharp occlusion boundaries but tend to lose high-frequency details. These methods, primarily designed for general 3D normal prediction, exhibit substantial errors when applied to relief images. Despite the advances of MonoRelief V2, a fidelity gap remains compared to the ground-truth, largely because the current model lacks a super-resolution mechanism to restore fine details lost during imaging.

\subsection{Applications}

\begin{figure*}[t] 
	\centering 
	{\tiny}	\includegraphics[width=6.7in]{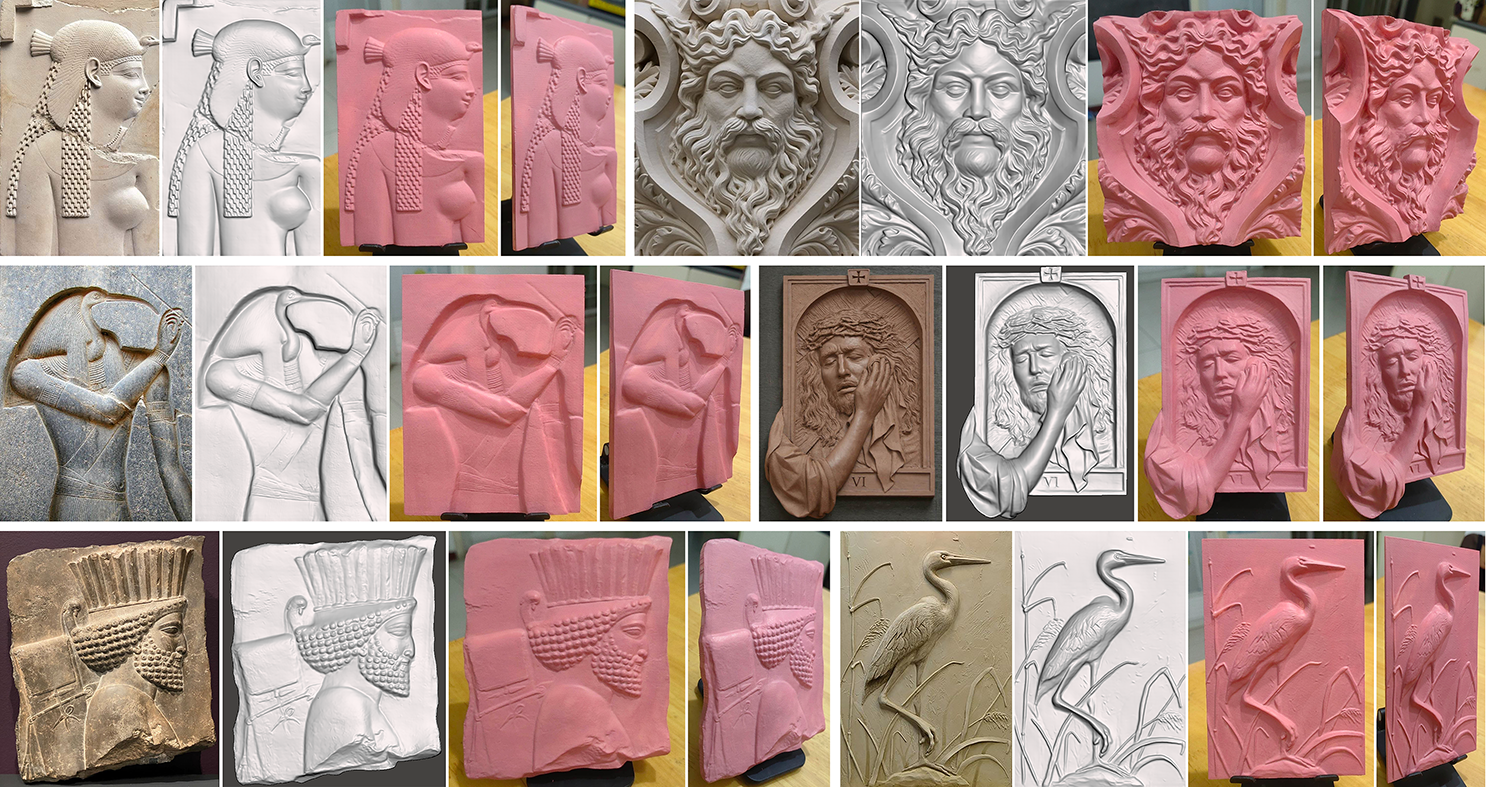} 
	\caption{Fabrication. Given an input image (left), whether sourced through photography, online resources, or generative image creation, MonoRelief V2 can recover the underlying relief geometry (middle), enabling subsequent production of physical reliefs through CNC machining or 3D printing (right).} 
	\label{fig13} 
\end{figure*}

Traditional relief creation relies heavily on manual craftsmanship, a process that demands specialized skills and considerable time investment. In contrast, generative 3D relief modeling dramatically enhance creative flexibility and production efficiency, established a new paradigm for cost-effective, personalized relief customization and heritage preservation. With its exceptional generalization capabilities, MonoRelief V2 can serve as a key technical foundation for such modeling. Building upon this, a potential integrated workflow is as follows: (1) Users first obtain design-aligned relief images via photography, online sourcing, or generative image creation; (2) MonoRelief V2 then recovers relief geometries from these single images; (3) Finally, the recovered reliefs are physically fabricated through laser engraving, CNC machining, or 3D printing. Fig. \ref{fig12} illustrates the results of MonoRelief V2 on photographs of real-world reliefs, AI-generated images, and software-rendered graphics respectively, while Fig. \ref{fig13} showcases examples of 3D-printed reliefs.
\subsection{Limitations}

\begin{figure*}[b] 
	\centering 
	{\tiny}	\includegraphics[width=6.7in]{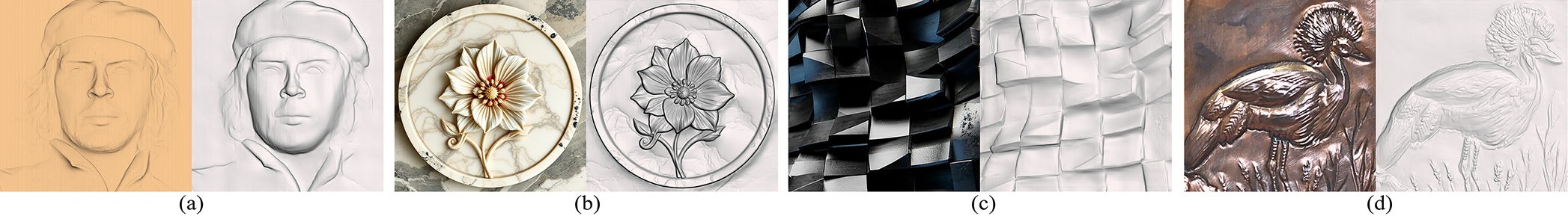} 
	\caption{Failure cases. (a) low-contrast shading. (b) texture pattern. (c) hard shadow. (d) strong specular highlights.} 
	\label{fig14} 
\end{figure*}

Despite achieving state-of-the-art performance in monocular relief recovery and demonstrating strong application potential, MonoRelief V2 currently faces several limitations. First, in the absence of image super-resolution capabilities, the quality of MonoRelief V2’s output is highly independent on the quality of the input image. It struggles to recover geometric details obscured by specular highlights or low-contrast shading, and cannot fully resolve geometric ambiguities caused by strong shadows and complex textures, as illustrated in Fig. \ref{fig14}. Second, due to the limited number of real-world data, the model has insufficient recognition of planar surfaces. Although planar reconstruction accuracy has improved significantly compared to V1, local planar regions-particularly those adjacent to steep gradients-still exhibit noticeable geometric noise. Third, constrained by the depth range of the training data, the model exhibits a marked increase in depth errors when processing relief scenes thicker than 400 pixels at 1024×1024 resolution.


\section{Conclusion}

This paper presents MonoRelief V2, the first end-to-end monocular relief recovery model trained on real relief data. To overcome the challenge of real data acquisition, we initially generated pseudo-real relief images with diverse shapes and materials using a text-to-image generative model, and synthesized corresponding depth pseudo-labels. Building upon this pseudo-real dataset, we obtained an initial image-to-depth model with strong generalization by fully fine-tuning DepthAnything V2 ~\cite{depthanything_v2} . To further improve prediction accuracy, we constructed a small-scale real-world dataset using multi-view 3D reconstruction and detail optimization techniques, followed by low-rank adaptation of the initial model. This progressive strategy significantly reduces depth errors while preserving generalization. As a result, MonoRelief V2 achieves state-of-the-art depth/normal estimation performance, surpassing both its predecessor V1 and other methods across multiple metrics. Although limitations remain, MonoRelief V2 provides crucial technical support for generative 3D relief modeling and shows strong potential for downstream applications. Future work will focus on expanding real relief datasets and incorporating additonal scene priors (e.g., background prior) into model training to further enhance the accuracy and robustness of relief recovery.

\section*{Acknowledgments}
We would like to thank the anonymous reviewers for their reviews and valuable suggestions. This work was supported in part by the National Natural Science Foundation of China (Grant No. 61772293, No. 62072274), the Joint Fund of the National Natural Science Foundation of China (U22A2033, U24A20219).This work was supported in part by the Major Innovation Project of Qilu University of Technology (Shandong Academy of Sciences)(No.2025ZDZX04), the National Natural Science Foundation of China (Grant No. 61772293, No. 62072274), the Joint Fund of the National Natural Science Foundation of China (U22A2033, U24A20219).

\clearpage
\newpage                                                   
\clearpage
\newpage
\FloatBarrier
\clearpage
\newpage
\bibliographystyle{IEEEtran}

%\bibliography{my_ref1}
{\small
	\bibliography{my_ref1}

% Generated by IEEEtran.bst, version: 1.14 (2015/08/26)
\begin{thebibliography}{10}
\providecommand{\url}[1]{#1}
\csname url@samestyle\endcsname
\providecommand{\newblock}{\relax}
\providecommand{\bibinfo}[2]{#2}
\providecommand{\BIBentrySTDinterwordspacing}{\spaceskip=0pt\relax}
\providecommand{\BIBentryALTinterwordstretchfactor}{4}
\providecommand{\BIBentryALTinterwordspacing}{\spaceskip=\fontdimen2\font plus
\BIBentryALTinterwordstretchfactor\fontdimen3\font minus
  \fontdimen4\font\relax}
\providecommand{\BIBforeignlanguage}[2]{{%
\expandafter\ifx\csname l@#1\endcsname\relax
\typeout{** WARNING: IEEEtran.bst: No hyphenation pattern has been}%
\typeout{** loaded for the language `#1'. Using the pattern for}%
\typeout{** the default language instead.}%
\else
\language=\csname l@#1\endcsname
\fi
#2}}
\providecommand{\BIBdecl}{\relax}
\BIBdecl

\bibitem{monorelief}
L.~Gao, Y.-W. Zhang \emph{et~al.}, ``Monorelief: Recovering 2.5d relief from a
  single image,'' \emph{IEEE Transactions on Visualization and Computer
  Graphics}, 2025, early Access.

\bibitem{stable_diffusion}
\BIBentryALTinterwordspacing
Stable diffusion. [Online]. Available: \url{https://stablediffusionweb.com/}
\BIBentrySTDinterwordspacing

\bibitem{flux}
\BIBentryALTinterwordspacing
Flux. [Online]. Available: \url{https://bfl.ai/models/flux-kontext}
\BIBentrySTDinterwordspacing

\bibitem{midjourney}
\BIBentryALTinterwordspacing
Midjourney. [Online]. Available: \url{https://www.midjourney.com/home}
\BIBentrySTDinterwordspacing

\bibitem{tripo3d}
\BIBentryALTinterwordspacing
Tripo3d. [Online]. Available: \url{https://studio.tripo3d.ai/home}
\BIBentrySTDinterwordspacing

\bibitem{meshy}
\BIBentryALTinterwordspacing
Meshy. [Online]. Available: \url{https://www.meshy.ai/}
\BIBentrySTDinterwordspacing

\bibitem{hunyuan3d}
\BIBentryALTinterwordspacing
Hunyuan3d. [Online]. Available: \url{https://3d.hunyuan.tencent.com/}
\BIBentrySTDinterwordspacing

\bibitem{omnidata}
A.~Eftekhar, A.~Sax, R.~Bachmann, J.~Malik, and A.~Zamir, ``Omnidata: A
  scalable pipeline for making multi-task mid-level vision datasets from 3d
  scans,'' in \emph{ICCV}, 2021.

\bibitem{depthanything_v2}
L.~Yang \emph{et~al.}, ``Depthanything v2,'' in \emph{NeurIPS}, 2024.

\bibitem{zeroshot_depth}
V.~Guizilini, I.~Vasiljevic, D.~Chen, R.~Ambrus, and A.~Gaidon, ``Towards
  zero-shot scale-aware monocular depth estimation,'' in \emph{ICCV}, 2023.

\bibitem{recover_3d_shape}
W.~Yin, J.~Zhang, O.~Wang, S.~Niklaus, L.~Mai, S.~Chen, and C.~Shen, ``Learning
  to recover 3d scene shape from a single image,'' in \emph{CVPR}, 2021.

\bibitem{metric3d}
W.~Yin, C.~Zhang, H.~Chen, Z.~Cai, G.~Yu, K.~Wang, X.~Chen, and C.~Shen,
  ``Metric3d: Towards zero-shot metric 3d prediction from a single image,'' in
  \emph{ICCV}, 2023.

\bibitem{zoedepth}
S.~F. Bhat, R.~Birkl, D.~Wofk, P.~Wonka, and M.~M{\"u}ller, ``Zoedepth:
  Zero-shot transfer by combining relative and metric depth,'' \emph{arXiv
  preprint arXiv:2302.12288}, 2023.

\bibitem{robust_monocular_depth}
R.~Ranftl, K.~Lasinger, D.~Hafner, K.~Schindler, and V.~Koltun, ``Towards
  robust monocular depth estimation: Mixing datasets for zero-shot
  cross-dataset transfer,'' \emph{TPAMI}, vol.~44, no.~3, 2022.

\bibitem{depth_anything}
L.~Yang, B.~Kang, Z.~Huang, X.~Xu, J.~Feng, and H.~Zhao, ``Depth anything:
  Unleashing the power of large-scale unlabeled data,'' in \emph{CVPR}, 2024.

\bibitem{diffusion_depth}
B.~Ke, A.~Obukhov, S.~Huang, N.~M. M{\"u}zger, R.~C. Daudt, and K.~Schindler,
  ``Repurposing diffusion-based image generators for monocular depth
  estimation,'' in \emph{CVPR}, 2024.

\bibitem{genpercept}
G.~Xu, Y.~Ge, M.~Liu, C.~Fan, K.~Xie, Z.~Zhao, H.~Chen, and C.~Shen, ``What
  matters when repurposing diffusion models for general dense perception
  tasks?'' in \emph{ICLR}, 2025.

\bibitem{lotus}
J.~He, H.~Li \emph{et~al.}, ``Lotus: Diffusion-based visual foundation model
  for high-quality dense prediction,'' in \emph{ICLR}, 2025.

\bibitem{latent_diffusion}
R.~Rombach, A.~Blattmann, D.~Lorenz, P.~Esser, and B.~Ommer, ``High-resolution
  image synthesis with latent diffusion models,'' in \emph{CVPR}, 2022.

\bibitem{metric3d_v2}
M.~Hu, W.~Yin, C.~Zhang \emph{et~al.}, ``Metric3d v2: A versatile monocular
  geometric foundation model for zero-shot metric depth and surface normal
  estimation,'' \emph{IEEE TPAMI}, 2024.

\bibitem{unidepth}
L.~Piccinelli, Y.-C. Yang, C.~Sakaridis, M.~Segu, S.~Li, and L.~V. Gool,
  ``Unidepth: Universal monocular metric depth estimation,'' in \emph{CVPR},
  2024.

\bibitem{unidepth_v2}
L.~Piccinelli, C.~Sakaridis, Y.-C. Yang, M.~Segu, S.~Li, W.~Abbeloos, and L.~V.
  Gool, ``Unidepthv2: Universal monocular metric depth estimation made
  simpler,'' \emph{arXiv preprint arXiv:2502.20110}, 2025.

\bibitem{depth_pro}
A.~Bochkovskii, A.~Delaunoy, H.~Germain, M.~Santos, Y.~Zhou, S.~R. Richter, and
  V.~Koltun, ``Depth pro: Sharp monocular metric depth in less than a second,''
  in \emph{ICLR}, 2025.

\bibitem{moge}
R.~Wang, S.~Xu, C.~Dai, J.~Xiang, Y.~Deng, X.~Tong, and J.~Yang, ``Moge:
  Unlocking accurate monocular geometry estimation for open-domain images with
  optimal training supervision,'' in \emph{CVPR}, 2025.

\bibitem{moge_v2}
R.~Wang, S.~Xu, Y.~Deng, Y.~Deng, J.~Xiang, Z.~Lv, G.~Sun, X.~Tong, and
  J.~Yang, ``Moge-2: Accurate monocular geometry with metric scale and sharp
  details,'' \emph{arXiv preprint arXiv:2507.02546}, 2025.

\bibitem{cross_task_consistency}
A.~R. Zamir, A.~Sax, N.~Cheerla, R.~Suri, Z.~Cao, J.~Malik, and L.~J. Guibas,
  ``Robust learning through cross-task consistency,'' in \emph{CVPR}, 2020.

\bibitem{aleatoric_uncertainty}
W.~Bae, I.~Budvytis, and R.~Cipolla, ``Estimating and exploiting the aleatoric
  uncertainty in surface normal estimation,'' in \emph{CVPR}, 2021, pp.
  13\,117--13\,126.

\bibitem{3d_corruptions}
O.~F. Kar, T.~Yeo, A.~Atanov, and A.~Zamir, ``3d common corruptions and data
  augmentation,'' in \emph{CVPR}, 2022.

\bibitem{rethink_inductive_biases}
W.~Bae and A.~J. Davison, ``Rethinking inductive biases for surface normal
  estimation,'' in \emph{CVPR}, 2024.

\bibitem{stablenormal}
C.~Ye, L.~Qiu, X.~Gu, Q.~Zuo, Y.~Wu, Z.~Dong, L.~Bo, Y.~Xiu, and X.~Han,
  ``Stablenormal: Reducing diffusion variance for stable and sharp normal,'' in
  \emph{SIGGRAPH Asia}, 2024.

\bibitem{finetune_diffusion}
G.~M. Garcia, K.~A. Zeid, C.~Schmidt, D.~de~Geus, A.~Hermans, and B.~Leibe,
  ``Fine-tuning image-conditional diffusion models is easier than you think,''
  in \emph{WACV}, 2025.

\bibitem{bilateral_normal}
X.~Cao, H.~Santo, B.~Shi, F.~Okura, and Y.~Matsushita, ``Bilateral normal
  integration,'' in \emph{ECCV}, 2022.

\bibitem{polycam}
\BIBentryALTinterwordspacing
Polycam. [Online]. Available: \url{https://poly.cam/}
\BIBentrySTDinterwordspacing

\bibitem{digital_basrelief}
T.~Weyrich, J.~Deng, C.~Barnes, S.~Rusinkiewicz, and A.~Finkelstein, ``Digital
  bas-relief from 3d scenes,'' \emph{ACM Transactions on Graphics}, vol.~26,
  no.~3, pp. 32--38, 2007.

\bibitem{basrelief_generation}
Y.-W. Zhang, Y.~Zhou, X.~Li, H.~Liu, and L.~Zhang, ``Bas-relief generation and
  shape editing through gradient-based mesh deformation,'' \emph{IEEE
  Transactions on Visualization and Computer Graphics}, vol.~21, no.~3, pp.
  328--338, 2015.

\bibitem{reliefnet}
Z.~Ji, W.~Feng, X.~Sun, F.~Qin, Y.~Wang, Y.-W. Zhang, and W.-y. Ma,
  ``Reliefnet: Fast bas-relief generation from 3d scenes,''
  \emph{Computer-Aided Design}, vol. 130, p. 102928, 2020.

\bibitem{portrait_relief}
Y.-W. Zhang, C.~Zhang, W.~Wang, Y.~Chen, Z.~Ji, and H.~Liu, ``Portrait relief
  modeling from a single image,'' \emph{IEEE Transactions on Visualization and
  Computer Graphics}, vol.~26, no.~8, pp. 2659--2670, 2020.

\bibitem{human_basrelief}
Z.~Yang, B.~Chen, Y.~Zheng, X.~Chen, and K.~Zhou, ``Human bas relief generation
  from a single photograph,'' \emph{IEEE Transactions on Visualization and
  Computer Graphics}, vol.~28, no.~12, pp. 4558--4569, 2025.

\bibitem{calligraphy_relief}
Y.-W. Zhang, J.~Wang, W.~Long, H.~Liu, C.~Zhang, and Y.~Chen, ``A fast solution
  for chinese calligraphy relief modeling from 2d handwriting image,''
  \emph{The Visual Computer}, vol.~36, no.~9, pp. 2241--2250, 2020.

\bibitem{flower_relief}
Y.-W. Zhang, J.~Wang, W.~Wang, Y.~Chen, H.~Liu, Z.~Ji, and C.~Zhang, ``Neural
  modelling of flower bas-relief from 2d line drawing,'' \emph{Computer
  Graphics Forum}, vol.~40, no.~6, pp. 288--303, 2021.

\bibitem{portrait_relief_v2}
Y.-W. Zhang, P.~Luo, H.~Zhou, Z.~Ji, H.~Liu, Y.~Chen, and C.~Zhang, ``Neural
  modeling of portrait bas-relief from a single image,'' \emph{IEEE
  Transactions on Visualization and Computer Graphics}, vol.~26, no.~8, pp.
  2659--2670, 2023.

\bibitem{multistyle_relief}
Y.-W. Zhang, H.~Yang, P.~Luo, Z.~Li, H.~Liu, Z.~Ji, and C.~Zhang, ``Modeling
  multi-style portrait relief from a single photograph,'' \emph{Graphical
  Models}, vol. 130, p. 101210, 2023.

\bibitem{mmrelief}
Y.-W. Zhang, Y.~Liu, H.~Yang, Y.~Chen, H.~Liu, Z.~Ji, and C.~Zhang, ``Mmrelief:
  Modeling multi-human relief from a single photograph,'' \emph{Computational
  Visual Media}, vol.~11, no.~3, pp. 531--548, 2025.

\bibitem{cgtrader}
\BIBentryALTinterwordspacing
Cgtrader. [Online]. Available: \url{https://www.cgtrader.com/}
\BIBentrySTDinterwordspacing

\bibitem{Zeel}
\BIBentryALTinterwordspacing
Zeel. [Online]. Available: \url{https://zeelproject.com/}
\BIBentrySTDinterwordspacing

\bibitem{Sketchfab}
\BIBentryALTinterwordspacing
Sketchfab. [Online]. Available: \url{https://sketchfab.com/}
\BIBentrySTDinterwordspacing

\bibitem{keyshot}
\BIBentryALTinterwordspacing
Keyshot. [Online]. Available: \url{https://www.keyshot.com/}
\BIBentrySTDinterwordspacing

\end{thebibliography}
}

\vspace{11pt}

\vfill

\clearpage
\onecolumn
\appendix

\subsection{Pseudo-Real Dataset}
\begin{figure}[h!] % 强制图片紧跟文字
	\centering
	\includegraphics[width=\textwidth]{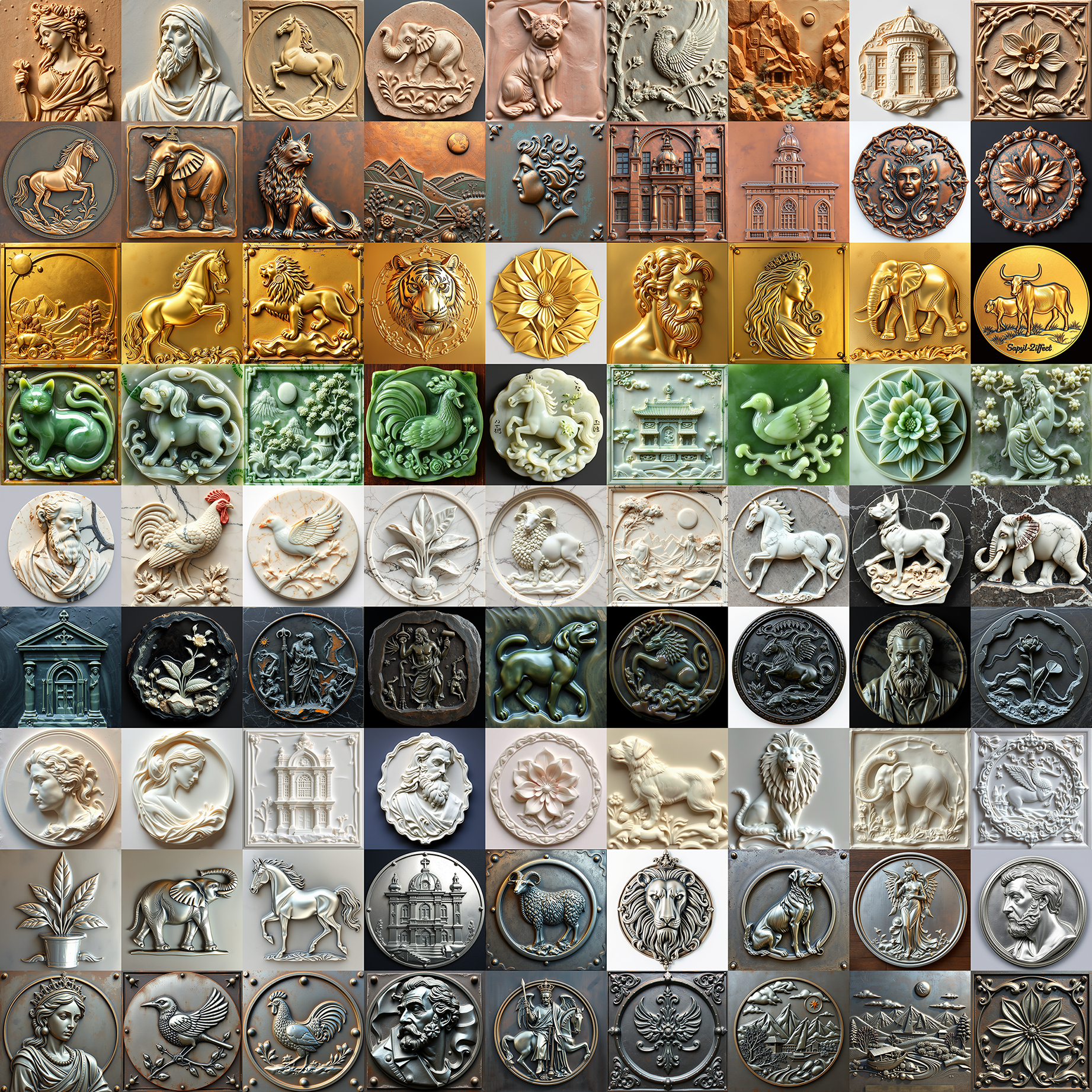} % 适应单栏宽度
	\caption{Pseudo-real relief images generated by Flux with diverse materials. From top to bottom: clay, copper, gold, jade, marble, obsidian, plaster, silver, and steel.}
	\label{fig15}
\end{figure}

The pseudo-real relief dataset consists of approximately 15,000 images generated by Flux at 1024×1024 resolution. It includes nine representative material types (clay, copper, gold, jade, marble, obsidian, plaster, silver, steel) and five common object categories (animals, human figures, plants, architecture, natural landscapes). Representative images are shown in Fig. \ref{fig15}.

\newpage
\subsection{Real-World Dataset}
\begin{figure}[h!]
	\centering
	\includegraphics[width=\textwidth]{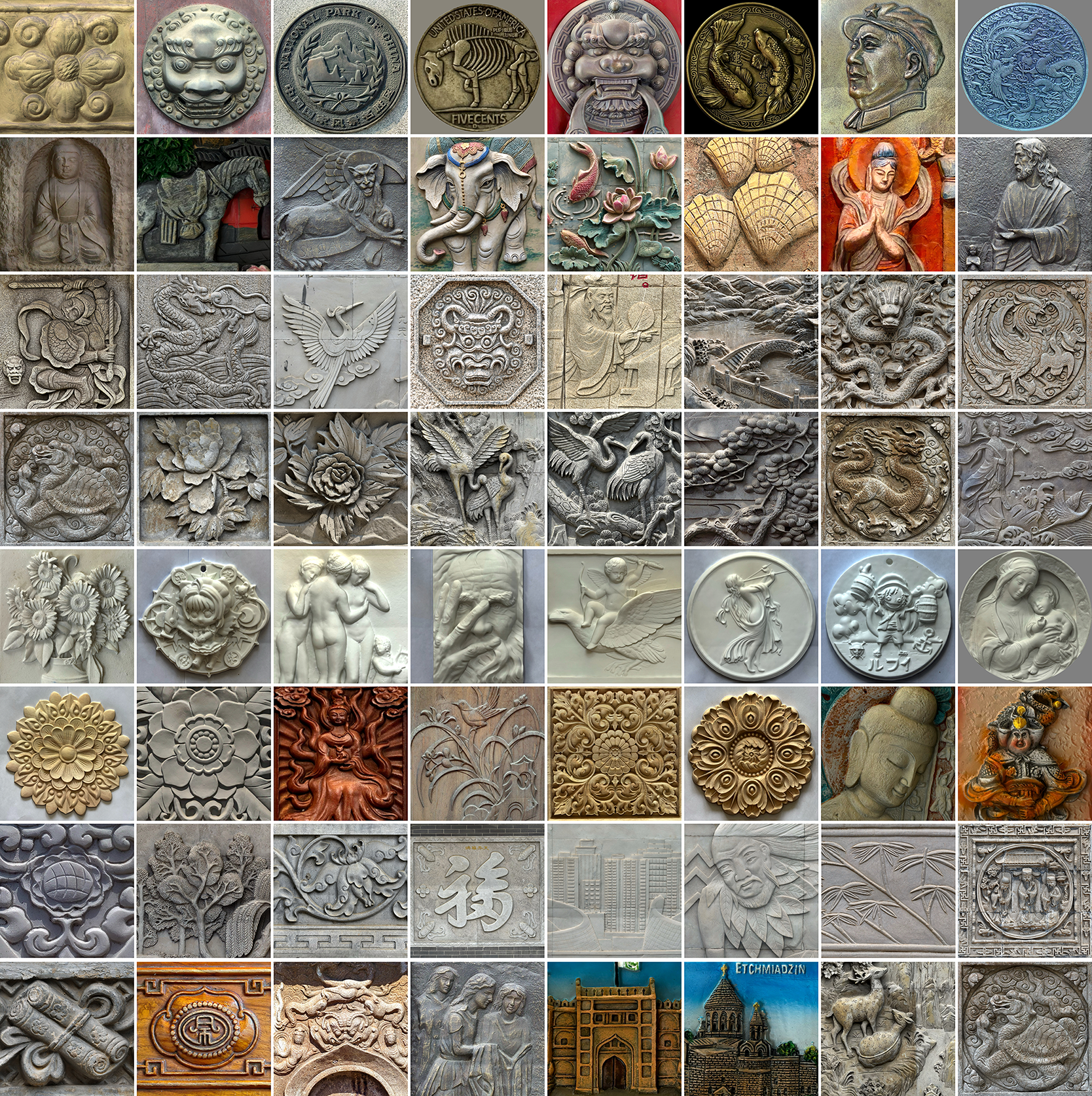}
	\caption{Real-world relief images.}
	\label{fig16}
\end{figure}

The real-world dataset comprises 800 samples sourced from physical reliefs crafted from materials such as plaster, brick, metal, wood. Each scene was constructed through multi-view 3D reconstruction and subsequently rendered in KeyShot at 1024×1024 resolution. Representative images are shown in Fig. \ref{fig16}.

\newpage
\subsection{Testing Benchmark}
\begin{figure}[h!]
	\centering
	\includegraphics[width=\textwidth]{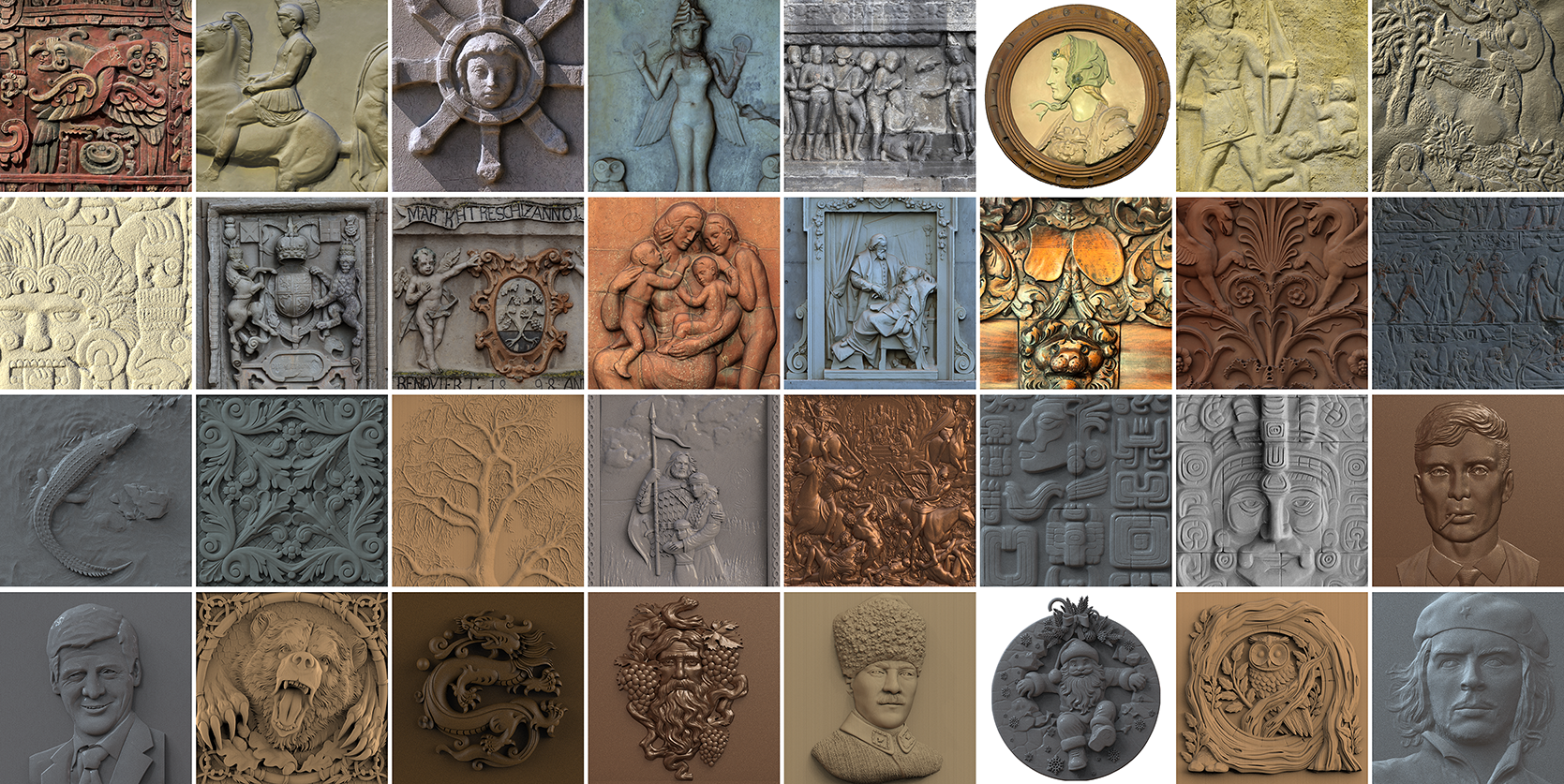}
	\caption{Real and synthetic relief images in testing benchmark.}
	\label{fig17}
\end{figure}

The testing benchmark integrates both synthetic and real data. The synthetic subset contains 252 samples, rendered in KeyShot from 36 hand-crafted 3D relief models, covering seven material variations (leather, concrete, high-gloss lacquer, bamboo texture, silver, marble, and porcelain). The real data subset includes 246 samples derived from 41 textured physical reliefs sourced from CGTrader, ZeeL, and Sketchfab, rendered under six distinct lighting conditions at 1024×1024 resolution. Benchmark samples are illustrated in Fig. \ref{fig17}.

\subsection{MonoRelief V1 versus V2}
\begin{figure}[h!]
	\centering
	\includegraphics[width=\textwidth]{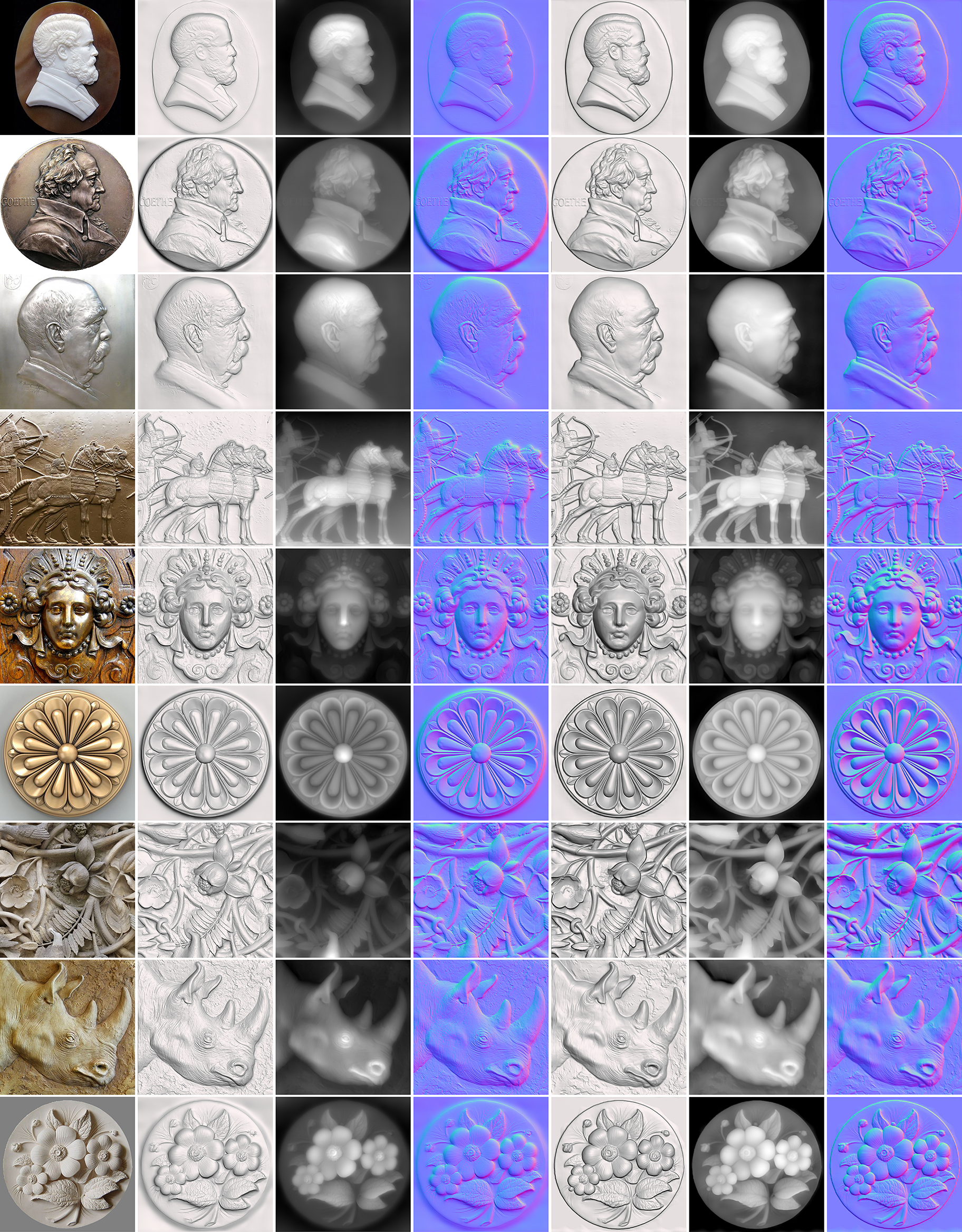}
	\caption{MonoRelief V1 (left) versus MonoRelief V2 (right)}
	\label{fig18}
\end{figure}

MonoRelief V2 demonstrates significant advancements over V1. First, it achieves efficiency gains through end-to-end prediction (100M parameters), enabling much faster inference. Second, its generalization capability is greatly enhanced by training on diverse real-world data, whereas V1 was limited to synthetic data. Third, reconstruction quality has been comprehensively improved: depth errors are significantly reduced, depth structures are more physically plausible, and local artifacts are greatly diminished. Comparative results are visualized in Fig. \ref{fig18}.

\subsection{Relief Generation}
\begin{figure}[h!]
	\centering
	\includegraphics[width=\textwidth]{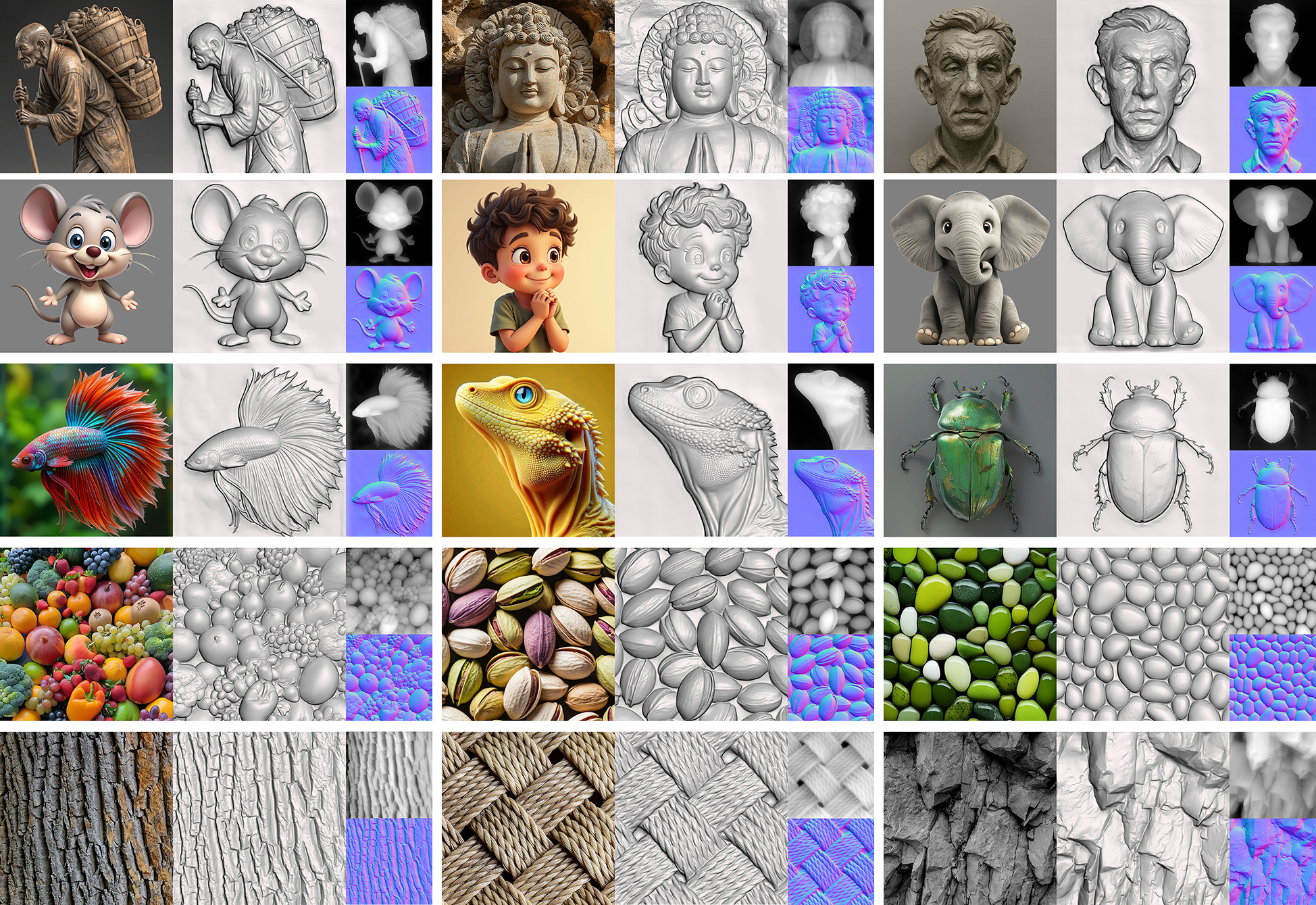}
	\caption{Relief generation on sculptures, cartoon, and objects with low-frequency textures.}
	\label{fig19}
\end{figure}

\begin{figure}[h!]
	\centering
	\includegraphics[width=\textwidth]{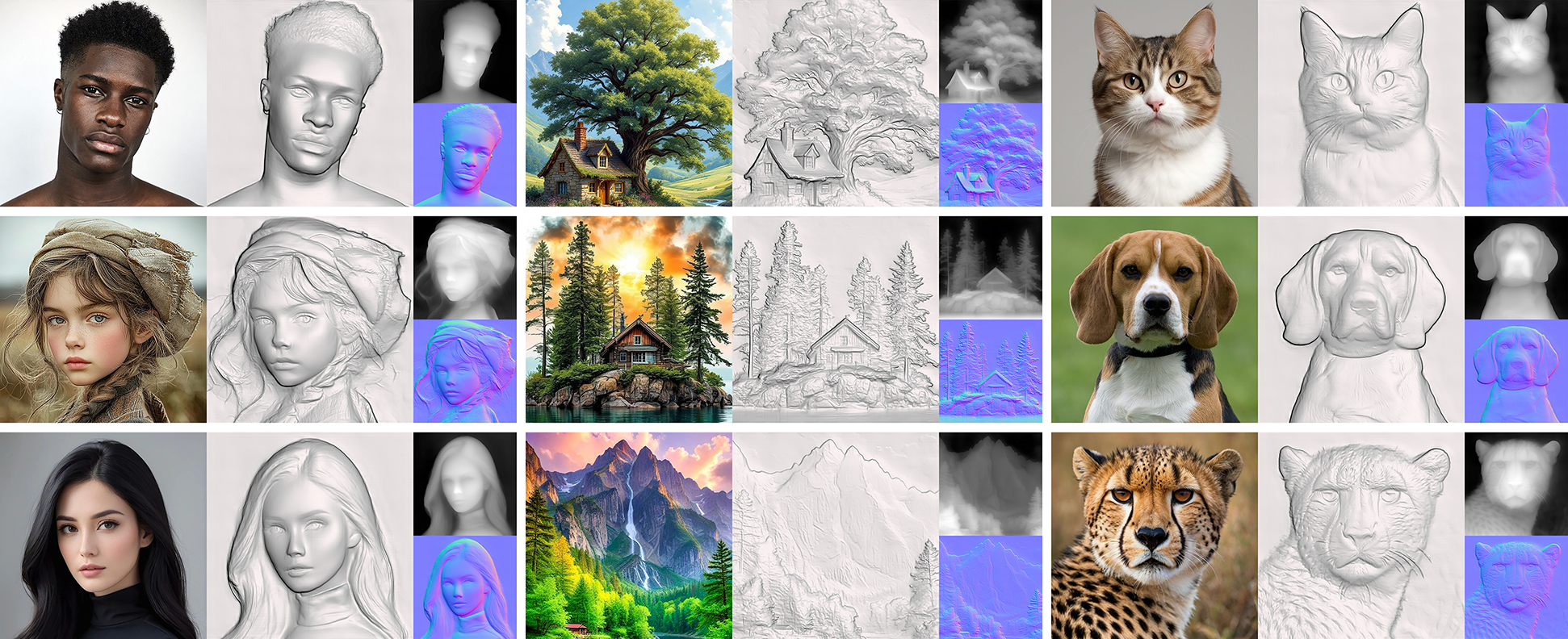}
	\caption{Relief generation on objects with low shading contrast or high-frequency textures.}
	\label{fig20}
\end{figure}

Beyond the relief domain, MonoRelief V2 can also generate reliefs from various source types, including sculptures, cartoon, and objects with high shading contrast and low-frequency texture variations, such as fish, lizards, beetles, fruits, nuts, stones, woven, and baked textures, as demonstrated in Fig. \ref{fig19}. Conversely, MonoRelief V2 exhibits limitations when handling natural landscapes, portraits, and animal images with low shading contrast or high-frequency texture variations. While fine details can be captured in such cases, significant geometric deviations from the ground truth occur, as illustrated in Fig. \ref{fig20}.

\end{document}